\documentclass{article}

\PassOptionsToPackage{numbers, compress}{natbib}

\usepackage[final]{neurips_2024}

\usepackage[utf8]{inputenc} 
\usepackage[T1]{fontenc}    
\usepackage[pagebackref,breaklinks,colorlinks,citecolor=green]{hyperref}
\usepackage{url}            
\usepackage{booktabs}       
\usepackage{amsfonts}       
\usepackage{nicefrac}       
\usepackage{microtype}      
\usepackage{xcolor}         
\usepackage{kotex}
\usepackage{bm}
\usepackage{multirow}
\usepackage{makecell}
\usepackage{graphicx}
\usepackage{mathtools}

\usepackage{amsmath}
\usepackage{amssymb}
\usepackage{xspace}
\usepackage{soul}
\usepackage{wrapfig}
\usepackage{subfig}
\usepackage{animate}
\usepackage{colortbl}

\makeatletter
\DeclareRobustCommand\onedot{\futurelet\@let@token\@onedot}
\def\@onedot{\ifx\@let@token.\else.\null\fi\xspace}

\def\rcell{\cellcolor[HTML]{ffb3b3}}
\def\ocell{\cellcolor[HTML]{ffd9b3}}
\def\ycell{\cellcolor[HTML]{ffffb3}}
\makeatother

\usepackage[capitalize]{cleveref}
\crefname{section}{Section}{Sections}
\Crefname{section}{Section}{Sections}
\crefname{table}{Table}{Tables}
\Crefname{table}{Table}{Tables}
\crefname{figure}{Figure}{Figures}
\Crefname{figure}{Figure}{Figures}
\crefname{equation}{Equation}{Equations}
\Crefname{equation}{Equation}{Equations}
\crefname{chapter}{Chapter}{Chapters}
\Crefname{chapter}{Chapter}{Chapters}
\crefname{algorithm}{Algorithm}{Algorithms}
\Crefname{algorithm}{Algorithm}{Algorithms}
\crefname{part}{Part}{Parts}
\Crefname{part}{Part}{Parts}

\newcommand{\psp}{\kern0.2ex}
\newcommand{\nsp}{\kern-0.1ex}

\newcommand{\tbf}[1]{\textbf{#1}}

\newcommand{\tworow}[2]{\begin{tabular}[c]{@{}c@{}}#1\vspace{-2pt}\\#2\end{tabular}}

\def\gG{{\mathcal{G}}}
\def\vx{{\bm{x}}}
\def\vc{{\bm{c}}}
\def\vp{{\bm{p}}}
\def\vm{{\bm{m}}}
\def\vr{{\bm{r}}}
\DeclarePairedDelimiter\floor{\lfloor}{\rfloor}
\newcommand{\train}{\mathcal{D}}

\newcommand{\Edited}[1]{#1}

\title{Fully Explicit Dynamic Gaussian Splatting}
\author{%
  Junoh Lee\textsuperscript{1}, Changyeon Won\textsuperscript{2}, Hyunjun Jung\textsuperscript{2}, Inhwan Bae\textsuperscript{2}, Hae-Gon Jeon\textsuperscript{1,2} \\
  \textsuperscript{1}School of Electrical Engineering and Computer Science 
  \textsuperscript{2}AI Graduate School \\
  Gwangju Institute of Science and Technology \\
  \texttt{\{juno,cywon1997,hyunjun.jung,inhwanbae\}@gm.gist.ac.kr, haegonj@gist.ac.kr} \\
}

\begin{document}

\maketitle

\begin{abstract}

3D Gaussian Splatting has shown fast and high-quality rendering results in static scenes by leveraging dense 3D prior and explicit representations. Unfortunately, the benefits of the prior and representation do not involve novel view synthesis for dynamic motions. Ironically, this is because the main barrier is the reliance on them, which requires increasing training and rendering times to account for dynamic motions. In this paper, we design a \Edited{Explicit 4D Gaussian Splatting(Ex4DGS)}. Our key idea is to firstly separate static and dynamic Gaussians during training, and to explicitly sample positions and rotations of the dynamic Gaussians at sparse timestamps. The sampled positions and rotations are then interpolated to represent both spatially and temporally continuous motions of objects in dynamic scenes as well as reducing computational cost. Additionally, we introduce a progressive training scheme and a point-backtracking technique that improves Ex4DGS's convergence. We initially train Ex4DGS using short timestamps and progressively extend timestamps, which makes it work well with a few point clouds. The point-backtracking is used to quantify the cumulative error of each Gaussian over time, enabling the detection and removal of erroneous Gaussians in dynamic scenes. Comprehensive experiments on various scenes demonstrate the state-of-the-art rendering quality from our method, achieving fast rendering of 62 fps on a single 2080Ti GPU.

\end{abstract}

\section{Introduction}
\label{sec:Intro}

The recent flood of video content has encouraged view synthesis techniques for highly engaging, visually rich content creation to maintain viewer interest. However, even short form videos require a huge time for both data pre-processing, computing frame-wise 3D point clouds, and training time in novel view synthesis for dynamic motions. Furthermore, these techniques must be considered for running on mobile devices which have a limited computing power and storage space. In this aspect, it is crucial to not only achieve photorealistic rendering results, but also reduce the computational cost related to storage, memory and rendering pipeline. To achieve this, the spatio-temporal representation should be explicit and efficient to handle the complexity of dynamic motions in videos.

Recent methods for dynamic view synthesis are typically based on Neural Radiance Fields (NeRF)~\cite{mildenhall2020nerf}, which uses implicit multi-layer perceptron (MLP) with a combination of 5-DoF spatial coordinates and the additional time axis~\cite{pumarola2021d,park2021nerfies,li2021neural,gao2021dynamic,park2021hypernerf,li2022neural,wu2022d}. The MLP-based methods have shown high-fidelity rendering quality. However, the inevitable cost of decoding the implicit representation makes rendering critically slow. Various methods try to reduce the cost by adapting explicit representation, such as voxel~\cite{liu2022devrf,li2022streaming} and matrix decomposition~\cite{cao2023hexplane,fridovich2023k}. Nevertheless, since NeRF-based approaches require per-pixel ray sampling and dense sampling for each ray, it is hard to achieve real-time and high-resolution renderings.

Meanwhile, a promising alternative, 3D Gaussian Splatting (3DGS)~\cite{kerbl20233d}, has emerged, which achieves photo-realistic rendering results with significantly faster training and rendering speeds. Unlike NeRF-based approaches, 3DGS exploits fully explicit point-based primitives of 3D and employs a rasterization-based rendering pipeline. Recent advances~\cite{wu20234dgaussians,li2023spacetime} attempt to extend 3DGS to 4D domain, handling the motion over time by storing the additional transformation information of 3D coordinates or 4D bases. However, these methods can only be trained under a restricted condition with dense point clouds. Moreover, these approaches borrow implicit representations for implementation, which lose the inherent advantage of 3DGS and make real-world applications more difficult. To become a more scalable model for real-world situations, it is important to be trained under more in-the-wild conditions (i.e., sparse point cloud) with more concise representation.

In this paper, we present \Edited{Explicit 4D Gaussian Splatting(Ex4DGS)}, a keyframe interpolation-based 4D Gaussian splatting method that works well in a fully explicit on-time domain. Our key idea is to apply interpolation techniques under temporal explicit representation to realize the scalable 3DGS model. Temporal interpolation is a widely used technique in computer graphics~\cite{zitnick2004high} that only stores keyframe information in video and determines smooth transitions for the rest of the frames. We select keyframes with sparse time intervals and save the additional temporal information at each keyframe which includes each Gaussian's position, rotation and opacity. This information is fully explicit; it is stored without any encoding process, and continuous motion is calculated to have smooth temporal transitions between adjacent keyframes. Here, we use a polynomial basis interpolator for position, a spherical interpolator for rotation, and a simplified Gaussian mixture model for opacity. 
Specifically, the polynomial basis of cubic Hermite spline (CHip)~\cite{bartels1995introduction} is used to effectively avoid overfitting or over-smoothing problems by spanning low-degree polynomials. 
For rotation, we introduce Spherical Linear Interpolation (Slerp)~\cite{Animating85ken} to do a linear transition over angles. Lastly, we introduce a simplified Gaussian mixture model, which allows temporal opacity to handle the appearance and disappearance of objects.

For further optimization, we reduce the computational cost by isolating dynamic points from static points, and only storing additional temporal information of dynamic points. Here, we aim for this separation to be possible without any additional inputs such as object masks~\cite{xu20234k4d}. To tackle this, we introduce motion-based triggers to distinguish static and dynamic points in a scene. 
We first initialize all Gaussian points in a scene to be static which are assumed to move linearly. During training, static points with large movements are automatically classified as dynamic points.
Next, we adopt a progressive training scheme to train our model even under sparse point cloud conditions. The progressive training, starting with a short duration and scaling up, can prevent falling into local minima by reducing the sudden appearance of objects. Lastly, an additional point backtracking technique is introduced to enhance rendering quality. Detecting redundant points in a dynamic scene is challenging because we need to consider all visible timestamps. To measure accumulated errors over time, we apply the point-backtracking technique that can track and prune high-error Gaussians in dynamic scenes.

The effectiveness of our method is validated through experiments on two major real-world video datasets: Neural 3D Video dataset~\cite{li2022neural} and Technicolor dataset~\cite{Sabater2017}. Experimental results demonstrate that our approach significantly improves rendering results even with sparse 3D point clouds. Furthermore, benefiting from the proposed optimization scheme, our model only requires low storage and memory size without any auxiliary modules or tricks to encode lengthy temporal information. Finally, our model achieves 62 fps on a single 2080Ti GPU on 300-frame scenes with a $1352\!\times\!1014$ resolution.

\vspace{-2mm}
\section{Related Works}
\label{sec:Rel}
\vspace{-2mm}
\subsection{Novel View Synthesis}
\label{subsec:sceneVR}
\vspace{-1mm}

Photorealistic novel view synthesis can be achieved using a set of densely sampled images through image-based rendering~\cite{levoy1996light, buehler2001unstructured}. While the dense sampling of views is limited by memory constraints and results in aliased rendering, novel view synthesis has advanced with the development of neural networks. The ability of neural networks to process implicit information in images enables novel view synthesis with a sparse set of observations. Prior works on constructing multi-plane images (MPI)\cite{zhou2018stereo, flynn2019deepview, srinivasan2019pushing} use aggregated pixel information from correspondences in sparsely sampled views. MPI representation may fail with wide angles between the camera and depth planes. This can be mitigated using geometric proxies like depth information \cite{kalantari2016learning, meshry2019neural}, plane-sweep volumes \cite{choi2019extreme, xu2019deep}, and meshes \cite{riegler2020free, riegler2021stable}. These methods risk unrealistic view synthesis with inaccurate proxy geometry. Joint optimization of proxy geometry \cite{hu2021worldsheet, chen2019learning, genova2018unsupervised, liu2019soft} can help, but direct mesh optimization often gets stuck in local minima due to poor loss landscape conditioning.

In recent years, continuous 3D representation through neural networks has received widespread attention. NeRF~\cite{mildenhall2020nerf} is the foundational work that started this trend. It implicitly learns the shape of an object as a density field, which makes optimization via gradient descent more tractable. The robustness of such geometric neural representations enables to reconstruct accurate geometry from a few given images~\cite{yu2021pixelnerf, kim2022infonerf, yang2023freenerf, jain2021putting}, large-scale geometry~\cite{tancik2022block, turki2022mega, barron2022mipnerf360, kaizhang2020, leegeometry, wang2023f2}, disentangled textures~\cite{yang2022neumesh, kellnhofer2021neural, yariv2020multiview}, and material properties~\cite{zhang2021nerfactor, zhang2021physg, yang2022ps, boss2022samurai, ye2023intrinsicnerf}. However, one major issue on these neural representations is also slow rendering speeds coming from volume renderers. To resolve this issue, works in \cite{sitzmann2021light,wang2022r2l} develop light field-inspired representations for single-pass pixel rendering. Both \cite{yu2021plenoctrees} and \cite{chen2022tensorf} introduce efficient voxel-based scene representations to improve the rendering speed. 
However, continuous representation inevitably combines with neural networks to implement its complex nature, and this limits rendering speed.
As an alternative, 3DGS~\cite{kerbl20233d} is designed to construct an explicit radiance field through rasterization, not requiring MLP-based inference. This method leverages anisotropic 3D Gaussians as the scene representation and proposes a differentiable rasterizer to render them by splatting onto the image plane for static scenes.
 
\vspace{-1.5mm}
\subsection{Dynamic Novel View Synthesis} 
\vspace{-0.5mm}
\label{subsec:sceneGP}

The time domain in dynamic scenes is typically parameterized with Plenoptic function~\cite{adelson1991plenoptic}. Classical image-based rendering~\cite{levoy1996light, buehler2001unstructured} and novel view synthesis methods using explicit geometry~\cite{riegler2020free, riegler2021stable} are restricted to a limited memory space~\cite{levoy1996light} when they extend to the time dimension. This is because they require additional optimization procedures for frame-by-frame dynamic view synthesis and storage for the parameterized space. Fueled by implicit representations, works in \cite{pumarola2021d, park2021nerfies, wu2022d, weng2022humannerf, li2021neural, li2022neural, gao2021dynamic} handle challenging tasks using neural volume rendering. They learn dynamic scenes by optimizing deformation and canonical fields for object motions~\cite{pumarola2021d, park2021nerfies, park2021hypernerf}, using human body priors~\cite{xu2021h, weng2022humannerf, alldieck2021imghum, loper2015smpl, jiang2023instantavatar, athar2022rignerf, bai2023high, peng2021animatable}, and decoupling dynamic parts~\cite{li2021neural, gao2021dynamic, xian2021space, zhang2021editable, tschernezki2021neuraldiff, li2023dynibar, jiang2022neuman, wu2022d, song2023nerfplayer}. These methods introduce additional neural fields or learnable variables to represent the time domain, taking more memory usage and rendering time as well. 

Temporally extended 3DGS has been considered as a feasible solution to dynamic novel view synthesis. A work in \cite{luiten2023dynamic} assigns parameters to 3DGS at each timestamp and imposes rigidity through a regularization. Another work in \cite{yang2023real} leverages Gaussian probability to model density changes over time to explicitly represent dynamic scenes. However, they require many primitives to capture complex temporal changes. Concurrently, works in \cite{wu20234dgaussians, yang2023deformable, huang2023sc, qian20233dgsavatar, hu2023gauhuman, lu20243d} utilize MLPs to represent the temporal changes. These methods inherit the drawbacks of dynamic neural representations, resulting in slower rendering speeds. The others in \cite{li2023spacetime, lin2023gaussian} explicitly parameterize the motion of dynamic 3D Gaussians to preserve the rendering speed of 3DGS by predicting their trajectory function. However, they only handle the motion as a continuous trajectory, and require multiple Gaussians for motions that disappear and reappear due to self-occlusion, increasing memory burden. 

In contrast, our key idea for dynamic 3D Gaussian uses keyframes to minimize primitives and to devise a progressive optimization to cope with scenarios where face disappearing/reappearing objects. Thanks to our schemes, we can improve rendering speed, memory efficiency, and achieve impressive performance for dynamic novel view synthesis.

\vspace{-2mm}
\section{Preliminary: 3D Gaussian Splatting}
\label{sec:Prelim}
\vspace{-1mm}

Our model starts from the point-based differentiable rasterization of 3DGS~\cite{kerbl20233d}. 3DGS uses three-dimensional Gaussian as geometry primitive, which is composed of position (mean) $\mu$, covariation $\bm{\Sigma}$, density $\sigma$ and color $\bm{c}$. The 3D Gaussian is referred to as follows:
\begin{equation}
\gG(\vx) = e^{-\frac{1}{2}(\vx-\mu)^\top\bm{\Sigma}^{-1}(\vx-\mu)}.
\end{equation}
We need to project the 3D Gaussian onto a 2D plane to render an image. In this process, the approximated graphics pipeline is used to render 2D Gaussians. The covariance matrix $\bm{\Sigma}'$ in camera coordinate is given as follows:
\begin{equation}
\bm{\Sigma}' = \bm{J}\,\bm{W}~\bm{\Sigma}~\bm{W}^{\!\top}\!\bm{J}^{\!\top},
\end{equation}
where $\bm{J}$ is the Jacobian of the affine approximation of the perspective projection and $\bm{W}$ is a viewing transformation. By skipping the third row and column of $\bm{\Sigma}'$, it is approximated to two-dimensional anisotropic Gaussian on the image plane.

The covariance is a positive semi-definite which can be decomposed into a scale $\bm{S}$ and a rotation $\bm{R}$ as:
\begin{equation}
\smash{\bm{\Sigma} = \bm{R}\psp\bm{S}\psp\bm{S}^\top\!\bm{R}^\top}.
\end{equation}

Spherical harmonics coefficients are used to represent view-dependent color changes as proposed in~\cite{fridovich2022plenoxels}. A rendered color from the Gaussian uses point-based $\alpha$ blending similar to NeRF’s volume rendering. For the interval between points along ray $\delta$ which can be obtained from $\mathcal{G}(\bm{x})$, the color of ray is
\begin{equation}
C = \sum_{i=1}^{N}T_{i}(1-e^{-\sigma_i \delta_i})\psp\vc_i,~T_i=e^{-\sum_{j=1}^{i-1}\sigma_j \delta_j},
\end{equation} 
where $N$ is the number of visible Gaussians along the ray, $i$ and $j$ denote the order of Gaussians by depth.

\begin{figure}[t]
\vspace{-1mm}
\centering
\includegraphics[width=\linewidth]{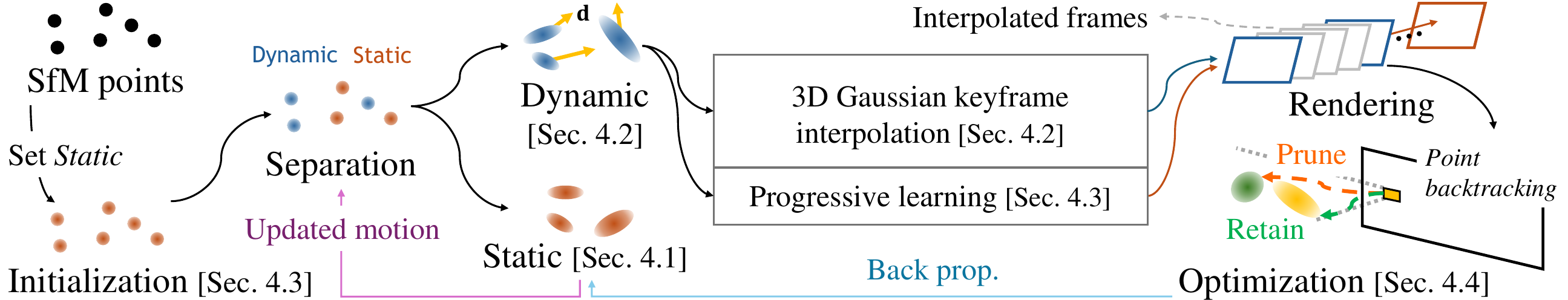}
\vspace{-6mm}
\caption{Overview of our method. We first initialize 3D Gaussians as static, modeling their motion linearly. During optimization, dynamic and static objects are separated based on the amount of predicted motion, and the 3D Gaussians between the selected keyframes are interpolated and rendered.}
\vspace{-2mm}
\label{fig:overview}
\end{figure}

\vspace{-2mm}
\section{Methodology}
\label{sub:Method}
\vspace{-1mm}
To achieve both memory efficiency and rendering capacity, our scheme is two-fold: (1) Keyframe-based interpolation to span position and rotation of Gaussians over time; (2) Classification of static and dynamic Gaussian. These are described in~\cref{subsec:StGS,subsec:DyGS}. After that, we introduce our progressive training scheme to handle a variety of running times in~\cref{sec:TempGS}, and deal with details of the optimization process of our method in~\cref{subsec:Opt}. The overview of our method is depicted in \cref{fig:overview}.

\vspace{-1.5mm}
\subsection{Static Gaussians}
\label{subsec:StGS}
\vspace{-0.5mm}
Static Gaussian $\mathcal{G}_s$ is modeled as the same as the 3DGS model except for its position. $\mathcal{G}_s$ changes the position linearly over time, which can be formulated with the position $\mu$ at time $t$ as below:
\begin{equation}
\smash{\mu(t) = \textbf{x} + t'\textbf{d}, t'=\frac{\,t\,}{l} \in [0, 1]}
\end{equation}
where $\textbf{x}$ is a pivot position of $\mathcal{G}_s$ and $\mathbf{d}$ is a vector representing the translation, and $l$ is the duration of a scene. We normalize $t$ with $l$ to prevent $\mathbf{d}$ from becoming too large.

\vspace{-1.5mm}
\subsection{Dynamic Gaussians}
\label{subsec:DyGS}
\begin{figure}[t]
\centering
\includegraphics[width=\linewidth]{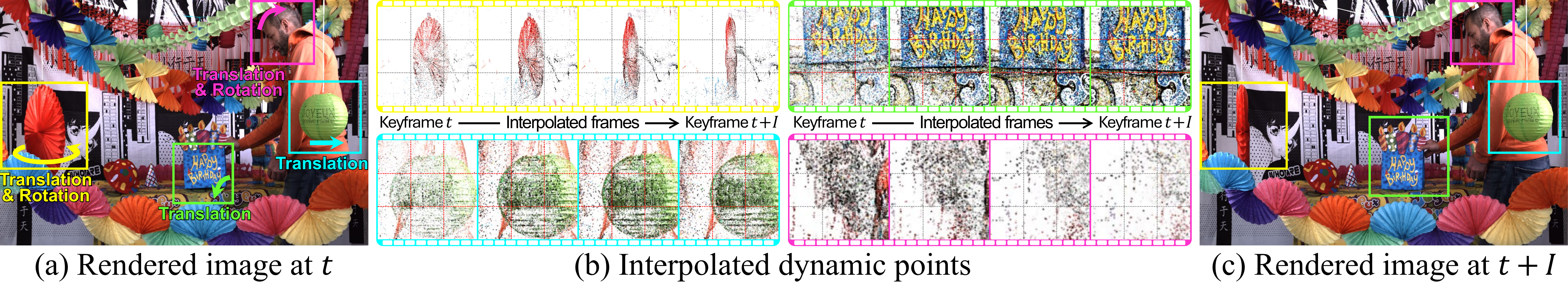}
\vspace{-6mm}
\caption{Effectiveness of our keyframe interpolation.}
\vspace{-4mm}
\label{fig:dynamic}
\end{figure}

\vspace{-0.5mm}
The dynamic Gaussian model is based on interpolations of keyframes, as visualized in \cref{fig:dynamic}. Specifically, the state of the dynamic Gaussian $\mathcal{G}_d$ at an intermediate timestamp is synthesized from adjacent keyframes. In this work, we assume that the keyframe interval is uniform for simplicity. The keyframe is defined as $\mathcal{K}=\{t\;|\;t=nI,n\in\mathbb{Z},t\in\mathcal{T}\}$ where $I$ is its interval and $\mathcal{T}$ is a set of timestamps. $\mathcal{G}_d$ acquires position $\mu$ and rotation in quaternion $r$ from keyframe information. We use different interpolators with different properties for smooth and continuous motion: CHip using polynomial bases is applied for positions, and a Slerp is used for rotations. We further adapt the Gaussian mixture model for temporal opacity to handle changes in the visibility of objects over time.

\vspace{-1.5mm}
\subsubsection{Cubic Hermite Interpolator for Temporal Position}
\label{subsec:CHip}
\vspace{-0.5mm}
CHip uses a third-degree polynomial. It is commonly used to model dynamic motions or shapes~\cite{bartels1995introduction}.
The interpolator function can be defined with third-degree polynomials and four variables: position and tangent vector of the start and end points. On the unit interval $[0,1]$, given a start point $p_0$ at $t=0$ and an end point $p_1$ at $t=1$ with start tangent $m_0$ at $t=0$ and an end tangent $m_1$ at $t=1$, $\textrm{CHip}$ can be defined as:

\begin{equation}
\begin{aligned}
\textrm{CHip}(\vp_0, \vm_0, \vp_1, \vm_1; t)~=~& (2t^3 - 3t^2 + 1)\vp_0 + (t^3 - 2t^2 + t)\vm_0 \\
&+ (-2t^3 + 3t^2)\vp_1 + (t^3 - t^2)\vm_1, \textit{~~~~~~~~where~~~} t \in [0, 1].
\end{aligned}
\end{equation}
Based on $\textrm{CHip}$, we compute the position $\mu$ of $\mathcal{G}_d$ at time $t$ as follows:
\begin{equation}
\begin{gathered}
\mu(t) = \textrm{CHip}(\vp_n, \vm_n, \vp_{n+1}, \vm_{n+1}; t'), \\
\textit{where~~} n = \smash{\floor*{\frac{t}{I}}},~~ t' = \frac{t - nI}{I},~~ \vm_n = \frac{\vp_{n+1} - \vp_{n-1}}{2I},~~ \vm_{n+1} = \frac{\vp_{n+2} - \vp_{n}}{2I},
\end{gathered}
\end{equation}
where $\bm{p}_n$ is a position of Gaussian at $n\mathrm{-th}$ keyframe. We use tangent values that is calculated using the position of two neighbor keyframes. This design can reduce additional requirements for storing tangent values at each keyframe, while still keeping the representational power for complex motion.

Other interpolators such as linear interpolation or piecewise cubic Hermite interpolating polynomial can be alternative choices. In this work, the cubic Hermit interpolator is selected because we can approximate the complex movements of points without any additional computational cost or storage.

\vspace{-1.5mm}
\subsubsection{Spherical Linear Interpolation for Temporal Rotation}
\label{subsec:Slerp}
\vspace{-0.5mm}
Slerp is typically used for interpolating rotations~\cite{Animating85ken} because linear interpolation causes a bias problem when it interpolates angular value.
On the unit interval $[0,1]$, given the unit vector $\bm{x}_0$ and $\bm{x}_1$ which represent rotations at $t=0$ and $t=1$ each, Slerp is defined as follows:
\begin{equation}
\mathrm{Slerp}(\vx_0, \vx_1; t) = \frac{\sin[(1-t)\Omega]}{\sin \Omega} \vx_0 + \frac{\sin[(t)\Omega]}{\sin \Omega} \vx_1 , \textit{~~~where~~~} t \in [0, 1] \text{~~and~} \cos\Omega = \vx_0 \cdot \vx_1.
\end{equation}
Slerp can be directly applied to quaternion rotations since it is independent of quaternions and dimensions. We thus have a rotation of intermediate frames in the quaternion of $\mathcal{G}_d$ at time $t$ without any modification as follows:
\begin{equation}
q(t) = \mathrm{Slerp}(\vr_n, \vr_{n+1}; t') \textit{~~~where~~~} n = \smash{\floor*{\frac{t}{I}}},~~ t' = \smash{\frac{t - nI}{I}},
\end{equation}
where $\bm{r}_n$ is the rotation of Gaussian at $n\mathrm{-th}$ keyframe.

\vspace{-0.5mm}
\subsubsection{Temporal Opacities}
\label{subsec:TempO}

\begin{wrapfigure}{r}{0.60\linewidth}
\vspace{-4mm}
\centering
\includegraphics[width=\linewidth]{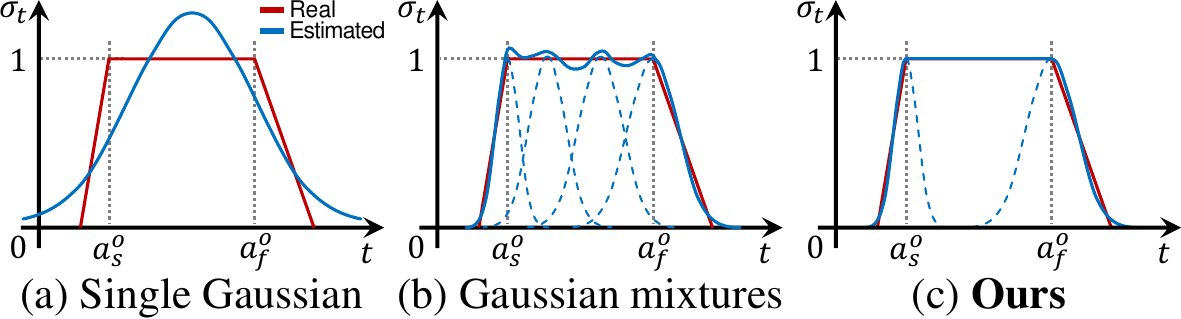}
\vspace{-6mm}
\caption{Comparison between the single Gaussian, Gaussian mixture, and our model for temporal opacity modeling.}
\vspace{-2mm}
\label{fig:temporal_opacity}
\end{wrapfigure}
Modeling the temporal opacity is important because it is directly related to appearing/disappearing objects. A straightforward model for temporal opacity is to directly use a single Gaussian. However, there is a limitation to model diverse cases only using the single Gaussian, such as sudden emerging/slowly vanishing objects and objects disappearing in videos, as illustrated in \cref{fig:temporal_opacity}. We introduce the Gaussian mixture model to handle these situations. Since using too many Gaussians is impractical, we approximate the model with two Gaussians.
We divide the temporal opacity into three cases: when an object appears, the object remains and the object disappears.
One Gaussian handles the appearance of the object and the other manages disappearance.
The interval between two Gaussians indicates the duration of the object when it is fully visible.

Let a Gaussian with a smaller mean value be $g_s^o$ and the other is $g_f^o$ where $a_s^o<a_f^o$. And, $a_s^o$, $b_s^o$, $a_f^o$ and $b_f^o$ are the mean and variance of $g_s^o$ and the mean and variance of $g_f^o$ each. The temporal opacity $\sigma_t$ at time $t$ is defined as follows:
\begin{equation}
\sigma_{t}(t) = 
\begin{cases}
e^{-\smash{\big(\frac{t-a^{o}_{s}}{b^{o}_{s}}\big)^2}}, & \textit{~~~for~~~} t < a^{o}_{s} \\
1, & \textit{~~~for~~~} a^{o}_{s} \leq t \leq a^{o}_{f} \\
e^{-\smash{\big(\frac{t-a^{o}_{f}}{b^{o}_{f}}\big)^2}}, & \textit{~~~for~~~} t > a^{o}_{f}~.
\end{cases}
\end{equation}
Using a single Gaussian may require multiple points to represent an object over a long duration.
In contrast, our model can handle both the short and long temporal opacity of an object using two Gaussians.

\begin{wrapfigure}{r}{0.30\linewidth}
\vspace{-4.5mm}
\centering
\includegraphics[width=\linewidth]{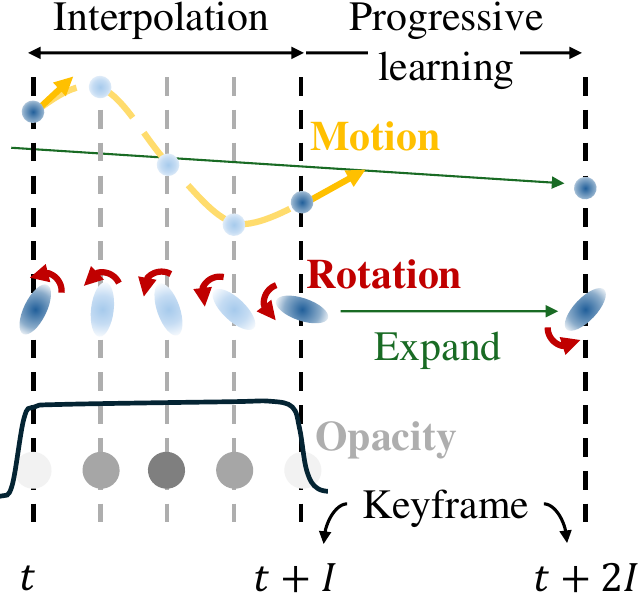}
\vspace{-6mm}
\caption{Progressive learning of dynamic Gaussians.
}
\vspace{-4mm}
\label{fig:interpolation}
\end{wrapfigure}

\vspace{-1.5mm}
\subsection{Training Scheme}
\label{sec:TempGS}
\vspace{-0.5mm}
\vspace{-1.5mm}
\paragraph{Progressive training scheme}\quad
Our goal is to minimize both memory and computational costs of the entire pipeline, including preprocessing, not just reducing the representation of 3DGS model in a dynamic scene. To this end, we adopt a progressive training scheme that allows to learn over a long duration using only a small amount of point clouds obtained from the first frame, which is illustrated in \cref{fig:interpolation}. To effectively handle objects moving or disappearing quickly, our model starts to learn a small part of an input video and gradually increases the video duration. The duration is incremented every specific step and made longer by the interval size.

\noindent\textbf{Expanding time duration}\quad
As the time duration increases, the number of keyframes in the dynamic Gaussian obviously increases. We estimate the position and rotation of the Gaussian by linear regression using the last $\rho$ frames when the number of keyframes increases so that the motion information of the previous frame can be shared with the next keyframe.

\noindent\textbf{Extracting dynamic points from static points}\quad
We want to separate dynamic points from static points without auxiliary information or supervision, such as masks or scribbles~\cite{li2023dynibar}. 
The separation is done based on the motion of the static points which is modeled to be movable, so we select the dynamic points based on the distance they moved. In particular, to avoid biased selection of distant points, we measure the motion in image space, normalizing the translation by the distance between points and the camera at the last timestamp. Therefore, if the distance to a point from the camera at the last timestamp is $\bm{\lambda}$, then the expression is $\frac{\lVert\mathbf{d}\rVert}{\lVert\bm{\lambda}\rVert^2}$. 
We sort points by the measured movement and convert the top-$\eta$ percent of points (in this work, $\eta=2$ is empirically set) into dynamic points. 
The position of the converted dynamic points is estimated at each keyframe using $\mathbf{x}$ and $\mathbf{d}$. The rotation is made to have the same value in all keyframes, and the opacity is initialized to be visible in all keyframes. We perform the extraction when we increase the duration or at specific iterations.

\vspace{-1.5mm}
\subsection{Optimization}
\label{subsec:Opt}
\vspace{-0.5mm}

\noindent\textbf{Point backtracking for pruning}\quad
Since it is difficult to filter out unnecessary dynamic points in a temporal context, we introduce a way to track errors in the image as points. Unlike contemporary works~\cite{chen2023gaussianeditor} that track points, we let our model track the value on a single backward pass. We use two measures, L1 distance and SSIM, whose formula is as follows:
\begin{equation}
\mathcal{E} = \frac{ \sum_{k} \big(\sigma_i \times \prod_{j=1}^{i-1}( 1-\sigma_j) \times q_k \big) }{ \sum_{k} \big(\sigma_i \times \prod_{j=1}^{i-1}( 1-\sigma_j)\big)},
\end{equation}
where $q_k$ is the measured error in image space, $k$ is a pixel index, and $i$ and $j$ are the order of Gaussian by depth which is visible at $k\mathrm{-th}$ pixel.
The accumulated error $\mathcal{E}_{total}$ is as follows:
\begin{equation}
\mathcal{E}_{total} = \frac{\sum_{v \in \train} \mathcal{E}_v}{\sum_{v \in \train}\mathbf{1}},
\end{equation}
where $\train$ is a set of training views.
We prune the points over $\mathcal{E}_{total}$ at every pre-defined step.

\noindent\textbf{Regularizations and losses}\quad
We use regularization for large motions on both static and dynamic points. The regularization minimizes $\lVert\mathbf{d}\rVert$ for static points and $\lVert\bm{p}_{n+1}-\bm{p}_n\rVert$ for dynamic points. The optimization process follows 3DGS, which uses differentiable rasterization based on gradient backpropagation. Both L1 loss and SSIM loss, which measure the error between a rendered image and its ground truth, are used.

\vspace{-1.5mm}
\subsection{Implementation Details}
\vspace{-0.5mm}
Our codebase is built upon \Edited{3DGS~\cite{kerbl20233d} and Mip-Splatting~\cite{Yu2024MipSplatting}} and uses almost its hyperparameters. 
For initialization, our experiments use only COLMAP~\cite{schonberger2016structure} point clouds from the first frame.
The time interval and initial duration are both set to $10$. We increment the duration by its interval every 400 iterations. Both static and dynamic regularization parameters are set to $0.0001$. We employ the RAdam optimizer~\cite{liu2019radam} for training.

\vspace{-2mm}
\section{Experiments}
\vspace{-1.5mm}
In this section, we conduct comprehensive experiments on two real-world datasets, Neural 3D Video~\cite{li2022neural} and Technicolor dataset~\cite{Sabater2017} in~\cref{subsec:N3V,subsec:Techni} each.
We follow a conventional evaluation protocol in~\cite{hyperreel,li2023spacetime}, which uses subsequences divided from whole videos.
We report PSNR, SSIM and LPIPS values. 
\Edited{To measure SSIM, we use \textit{scikit\_image} library. Here, SSIM$_1$ and SSIM$_2$ are computed using \textit{data\_range} value of $1$ and $2$, respectively.} We also measure frame-per-second (FPS) for rendering speed, and training time including preprocessing time.
To compare the robustness of our method according to initial point clouds, we additionally test contemporary works on sparse point cloud initialization, which uses only the point cloud of the first frame in~\cref{subsec:N3V,subsec:Techni}.
We also visualize the separation of static and dynamic points to show that our model can successfully distinguish them in~\cref{subsec:sepDS}.
The ablation study shows the effectiveness of each component in our model in~\cref{subsec:ablation}.

\newcommand{\NA}{\textcolor{lightgray}{\footnotesize N/A}}
\begin{table}[t]
\caption{Comparison of ours with the comparison methods on N3V dataset~\cite{li2022neural}. Training time: Both preprocessing and the accumulated time of all subsequent training phases. Both the training time and FPS are measured under the same machine with an NVIDIA 4090 GPU for strictly fair comparisons. {\small\textdagger}: STG is done with an H100 GPU machine due to the memory issue. {\small\textdaggerdbl}: Trained using a dataset split into 150 frames.}
\vspace{1mm}
\centering
\resizebox{\linewidth}{!}{%
\begin{tabular}{ccccccccccc}
\toprule
\multirow{2}{*}{Model\vspace{-15pt}} & \multicolumn{7}{c}{PSNR (dB)} & ~~~~MB~~~~ & Frame/s & Hours \\ \cmidrule(lr){2-8} \cmidrule(lr){9-9} \cmidrule(lr){10-10} \cmidrule(lr){11-11}
 & \tworow{Coffee}{Martini} & \tworow{Cook}{Spinach} & \tworow{\!\!\!\!\!Cut\,Roasted\!\!\!\!\!}{Beef} & \tworow{Flame}{Salmon} & \tworow{Flame}{~Steak~} & \tworow{Sear}{~Steak~} & Average & Size & FPS & \tworow{\!Training\!}{time}\\ \midrule
NeRFPlayer~\cite{song2023nerfplayer}       & 31.53 & 30.56 & 29.35 & 31.65 & 31.93 & 29.13 & 30.69 & 5130 & 0.05 & 6 \\
HyperReel~\cite{hyperreel}        & 28.37 & 32.30 & 32.92 & 28.26 & 32.20 & 32.57 & 31.10 & 360 & 2 & 9 \\
Neural Volumes~\cite{lombardi2019neural}                & \NA   & \NA   & \NA   & 22.80 & \NA   & \NA   & 22.80 & \NA & \NA  & \NA \\ 
LLFF~\cite{mildenhall2019llff}                          & \NA   & \NA   & \NA   & 23.24 & \NA   & \NA   & 23.24 & \NA & \NA  & \NA \\ 
DyNeRF~\cite{li2022neural}                              & \NA   & \NA   & \NA   & 29.58 & \NA   & \NA   & 29.58 & 28  & 0.015& 1344\\ 
HexPlane~\cite{cao2023hexplane}                         & \NA   & 32.04 & 32.55 & 29.47 & 32.08 & 32.39 & 31.71 & 200 & \NA  & 12  \\ 
K-Planes~\cite{fridovich2023k}                          & 29.99 & 32.60 & 31.82 & 30.44 & 32.38 & 32.52 & 31.63 & 311 & 0.3  & 1.8 \\ 
MixVoxels-L~\cite{wang2023mixed}                        & 29.63 & 32.25 & 32.40 & 29.81 & 31.83 & 32.10 & 31.34 & 500 & 37.7 & 1.3 \\ 
MixVoxels-X~\cite{wang2023mixed}                        & 30.39 & 32.31 & 32.63 & 30.60 & 32.10 & 32.33 & 31.73 & 500 & 4.6  & \NA \\ 
Im4D~\cite{lin2023im4d}                                 & \NA   & \NA   & 32.58 & \NA   & \NA   & \NA   & 32.58 & \NA & \NA & \NA \\ 
4K4D~\cite{xu20234k4d}                                  & \NA   & \NA   & 32.86 & \NA   & \NA   & \NA   & 32.86 & \NA & 110 & \NA \\ \cmidrule(lr){1-11}
\multicolumn{11}{c}{Dense COLMAP point cloud input}   \\
STG\textsuperscript{\textdaggerdbl}~\cite{li2023spacetime} & 28.41 & 32.62 & 32.53 & 28.61 & 33.30 & 33.40 & 31.48 & 107 & 88.5    & 5.2\textsuperscript{\textdagger}\!\! \\
4DGS~\cite{yang2023real}                                & 28.33 & 32.93 & 33.85 & 29.38 & 34.03 & 33.51 & 32.01 & 6270& 71.4    & 5.5 \\
4DGaussians~\cite{wu20234dgaussians}                    & 27.34 & 32.46 & 32.90 & 29.20 & 32.51 & 32.49 & 31.15 & 34  & 136.9    & 1.7 \\ \cmidrule(lr){1-11}
\multicolumn{11}{c}{Sparse COLMAP point cloud input}   \\
STG\textsuperscript{\textdaggerdbl}~\cite{li2023spacetime} & \ycell 27.71 & 31.83 & 31.43 & \ycell 28.06 & \ycell 32.17 & \ycell 32.67 & \ycell 30.64 & \ocell 109 & \ycell 101.0    &  \ocell 1.3\textsuperscript{\textdagger} \\
4DGS~\cite{yang2023real}                                & 26.51 & \ycell 32.11 & \ycell 31.74 & 26.93 & 31.44 & 32.42 & 30.19 & 6057& 72.0    & 4.2 \\
4DGaussians~\cite{wu20234dgaussians}                    & 26.69 & 31.89 & 25.88 & 27.54 & 28.07 & 31.73 & 28.63 & \rcell 34  & \rcell146.6    & \ycell 1.5 \\
3DGStream~\cite{sun20243dgstream}                                       & \ocell 27.75\ocell & \rcell 33.31 & \ocell 33.21 & \ocell 28.42 & \rcell 34.30 & \ocell 33.01 & \ocell 31.67 & 1200 & -    & - \\
\tbf{Ours}                                              & \rcell 28.79 & \ocell 33.23 & \rcell 33.73 & \rcell 29.29 & \ocell 33.91 & \rcell 33.69 & \rcell 32.11 & \ycell 115 & \ocell 120.6 & \rcell 0.6 \\
\bottomrule
\end{tabular}%
}
\vspace{-2mm}
\label{tab:N3V}
\end{table}

\vspace{-1.5mm}
\subsection{Neural 3D Video Dataset}
\label{subsec:N3V}
\vspace{-0.5mm}
Neural 3D Video dataset~\cite{li2022neural} provides six sets of multi-view indoor videos, captured with a range of 18 to 21 cameras with a 2704$\times$2028 resolution and 300 frames. Following the conventional evaluation protocol, both training and evaluation procedures are performed at half the original resolution, 
and the center camera is held out as a novel view for evaluation.
For a fair comparison, we train all models for all 300 frames including concurrent works, except for STG~\cite{li2023spacetime}, NeRFPlayer~\cite{song2023nerfplayer} and HyperReel~\cite{hyperreel}.
For NeRFPlayer and HyperReel, we directly borrow the results from~\cite{song2023nerfplayer,hyperreel}.
For STG, it is not possible to train for all 300 frames due to a GPU memory issue, so we report the results for only 150 frames, which is the maximum duration running on a single NVIDIA H100 80GB GPU.

\begin{figure}[t]
\centering
\includegraphics[width=\linewidth]{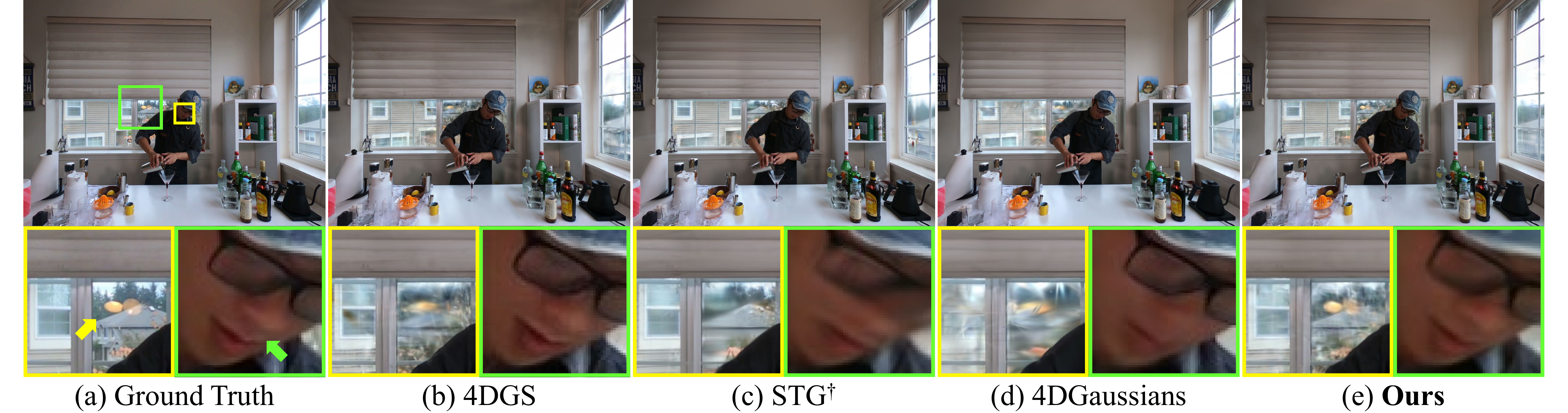}
\vspace{-6mm}
\caption{Comparison of our Ex4DGS with other the state-of-the-art dynamic Gaussian splatting methods on Neural 3D Video~\cite{li2022neural} dataset.}
\vspace{-1mm}
\label{fig:n3v}
\end{figure}

As shown in~\cref{tab:N3V}, our model outperforms most of the contemporary models while maintaining the low computational cost. 
The example is displayed in \cref{fig:n3v}, which shows that our model produces high-quality rendering results over the comparison methods.

\textbf{Comparison on sparse conditions}\quad
We also carry out an experiment to check if concurrent methods work well with sparse point cloud initialization, which uses only it for the first frame.
We report the result in \cref{tab:N3V}. Interestingly, all the concurrent methods yield unsatisfactory results because motions in videos are learned by relying on the point clouds, not temporal changes of objects in the training phase.
This implies that they require well-reconstructed 3D point clouds for proper initialization, while our method is free from the initial condition.

\begin{wraptable}{r}{0.53\linewidth}
\vspace{-4.5mm}
\centering
\caption{Comparison results on\,the\,Technicolor\,dataset \cite{Sabater2017}. {\small\textdagger}: Trained with sparse point cloud input.}
\vspace{-1mm}
\resizebox{\linewidth}{!}{%
\begin{tabular}{ccccc}
\toprule
Model & ~PSNR~ & ~SSIM$_1$\!~ & ~SSIM$_2$\!~ & ~LPIPS~ \\ \midrule 
DyNeRF~\cite{li2022neural}                              & 31.80 &  \NA  & \ycell 0.958  & 0.142 \\
HyperReel~\cite{hyperreel}                              & \ycell 32.73 & \ycell 0.906 &  \NA   & \ycell 0.109 \\
4DGS~\cite{yang2023real}                              & 29.54 & 0.873 & 0.937 & 0.149 \\
4DGaussians~\cite{wu20234dgaussians}                              & 30.79 & 0.843 & 0.921 & 0.178 \\
STG\textsuperscript{\textdagger}~\cite{li2023spacetime} & \ocell 33.23 & \ocell 0.912 & \ocell 0.960 & \rcell 0.085 \\
\textbf{Ours}                                           & \rcell 33.62 & \rcell 0.916 & \rcell 0.962 & \ocell 0.088 \\
\bottomrule
\end{tabular}%
}
\vspace{-3mm}
\label{tab:techni}
\end{wraptable}

\vspace{-1.5mm}
\subsection{Technicolor Dataset}
\label{subsec:Techni}
\vspace{-0.5mm}
Technicolor light field dataset encompasses video recordings captured using a 4$\times$4 camera array, wherein each camera is synchronized temporally, and the spatial resolution is 2048$\times$1088. Adhering to the methodology introduced in HyperReel~\cite{hyperreel}, we reserve the camera positioned at the intersection of the second row and second column for evaluation purposes. Evaluation is conducted on five distinct scenes (Birthday, Fabien, Painter, Theater, Trains) using their original full resolution.
We retrain STG~\cite{li2023spacetime} using the COLMAP point cloud from the first frame, instead of the point cloud from every frame, for strictly fair comparison.

As shown in \cref{tab:techni}, Ex4DGS is comparable with the second-best model in the sparse COLMAP scenario. Although the Technicolor dataset contains various colorful objects, our model successfully synthesizes the novel view without dense prior or additional parameters. The reason why STG shows the impressive performance is that Technicolor dataset does not have rapid movements.

\begin{figure}[t]
\centering
\includegraphics[width=\linewidth]{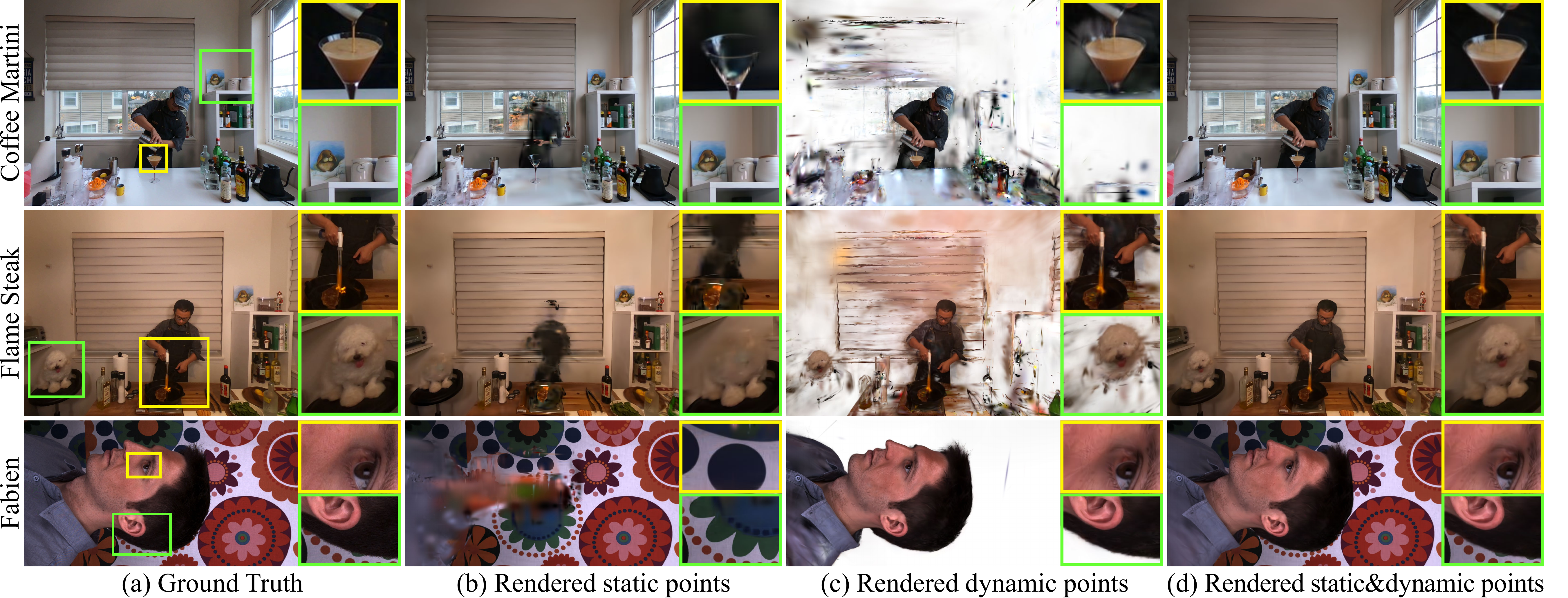}
\vspace{-6mm}
\caption{Visualization of our static and dynamic point separation on Coffee Martini, Flame Steak and Fabien scene in Neural 3D Video~\cite{li2022neural} and Technicolor~\cite{Sabater2017} datasets.}
\vspace{-2mm}
\label{fig:SDsep}
\end{figure}

\vspace{-1.5mm}
\subsection{Separation of Dynamic and Static Points}
\label{subsec:sepDS}
\vspace{-0.5mm}

Ex4DGS has a capability to separate static and dynamic points during the learning process. To check how well they are separated, we render them individually. \cref{fig:SDsep} shows the separation result. The static and dynamic points are rendered on both the Neural 3D Video and Technicolor datasets. The results demonstrate that the dynamic points are successfully separated from the static points, even if they are trained in an unsupervised manner.
As a result, view-dependent color-changing or reflective objects are also identified as dynamic parts. Furthermore, in Coffee Martini scene, Ex4DGS demonstrates the ability to detect dynamic fluid in the transparent glasses. It is also worth highlighting that the same object can have static and dynamic components, as shown in the dog's legs and head being classified as distinct points in the Flame Steak scene.

\begin{wraptable}{r}{0.59\linewidth}
\vspace{-4.5mm}
\caption{Ablation studies of the proposed methods.}
\vspace{-1.5mm}
\centering
\resizebox{\linewidth}{!}{%
\begin{tabular}{ccccc}
\toprule
Method & \,PSNR\, & \,SSIM$_1$\!\, & \,LPIPS\, & \!\!Size(MB)\!\! \\ \midrule
w/ Linear position                   &        31.12 &        0.9385 &        0.0524 &        204 \\
w/o Temporal opacity                 & \ocell 31.42 & \ycell 0.9394 & \ocell 0.0521 &        186 \\
w/ Linear rotation                   &        31.26 &        0.9392 &        0.0525 & \ycell 148 \\
w/o Progressive growing              &        31.02 &        0.9389 &        0.0550 &        168 \\
w/ Linear position\&rotation         &        31.32 & \ycell 0.9394 & \ocell 0.0521 &        172 \\
w/o Regularization                   &        31.37 & \ocell 0.9395 & \ycell 0.0522 &        174 \\ 
\!\!w/o Dynamic point extraction\!\! &        28.58 &        0.9280 &        0.0756 & \rcell 58 \\
w/o Point backtracking               & \ycell 31.40 & \ocell 0.9394 &        0.0529 &        169 \\ \hline
\tbf{Ours}                           & \rcell 32.11 & \rcell 0.9422 & \rcell 0.0478 & \ocell 115 \\
\bottomrule
\end{tabular}%
}
\vspace{-5mm}
\label{tab:abl}
\end{wraptable}

\vspace{-1.5mm}
\subsection{Ablation Studies}
\label{subsec:ablation}
\vspace{-0.5mm}

We conduct an extensive ablation study to check the effectiveness of the proposed technical components in~\cref{tab:abl}. 

We first examine the effectiveness of our interpolation method by changing them into linear models. The results show that linear modeling of the position and rotation reduces the quality of rendering. Interestingly, using different types of interpolations further exacerbates the performance. If an equal level of polynomial bases are not assigned to both attributes, one of them falls short of the representational capacity, resulting in overfitting or over-smoothing. 

We also show how our dynamic point extraction affects the rendering quality. As expected, complex motions can only be handled with dynamic point modeling. We then evaluate the efficacy of our temporal opacity modeling, and observe the performance degradation when no temporal opacity changes. Points can only disappear by minimizing the size and hiding back to other Gaussian points, making them act as flutters without being removed properly. 

We then check the effectiveness of our progressive growing strategy. Without this strategy, the optimization gets stuck in local minima. This is due to our approach using only the point cloud from the first frame, which results in a misalignment with their corresponding objects of future frames. 

We evaluate our regularization terms for the temporal dynamics of both static and dynamic points within a scene. As expected, incorporating the additional regularization term into the learning process makes the dynamic scene representations better. This benefit comes from the reduction of accumulated motion errors, preventing the points from moving excessively and locating them at correct positions. 

Lastly, we examine the effectiveness of our point backtracking approach for pruning step. As expected, the correct removal of the misplaced points mitigates the errors and leads to the best result.

\vspace{-1.5mm}
\section{Conclusion}
\label{conclusion}
\vspace{-0.5mm}
We have proposed a novel parameterization of dynamic 3DGS by explicitly modeling 3D Gaussians' motions. To achieve this, we initially set keyframes and predict their position and rotation changes. Primitive parameters of the 3D Gaussians between keyframes are then interpolated. 
Our strategy for learning dynamic motions enables us to decouple static and dynamic parts of scenes, opening up more intuitive and interpretable representation in 4D novel view synthesis.

\noindent\textbf{Limitations}\quad 
Although we achieve a memory-efficient explicit representation of dynamic scenes, two challenges remain. First, our reconstruction can get stuck in local minima for newly appeared objects that are not initialized with 3D points and have no relevant 3D Gaussians in neighboring frames. This issue could be mitigated by initializing new 3D points with an additional geometric prior such as depth information. Second, as 3DGS suffers from scale ambiguity, training on monocular videos is challenging. This is because every 3D Gaussians are treated as dynamic due to the lack of accurate geometric clues for objects at each timestamp. This challenge can be addressed by incorporating an additional semantic cue information like object mask and optical flow, which account for objects' motions more explicitly.

\clearpage

\appendix

\section{Overview}
\label{appendix_overview}
Within the appendix, we provide additional experiments in~\cref{appendix_additional_exp}, additional comparisons in~\cref{appendix_more_comparison} and per-scene breakdown of quantitative comparisons in~\cref{appendix_deatil_results}.

\section{Additional Experiments}
\label{appendix_additional_exp}
In this section, we conduct experiments to further illustrate the behavior of Ex4DGS. In~\cref{appendix_wo_handling_color}, we present an experiment where changes in color alone are not treated as dynamic points. In~\cref{appendix_repeative}, we demonstrate how Ex4DGS behaves when objects reappear. In~\cref{appendix_keyframe_sel} and~\cref{appendix_diff_conversion}, we present additional ablation studies on various keyframe interval selection and dynamic point conversion rates.

\subsection{Without Handling Color Components}
\label{appendix_wo_handling_color}
\begin{figure*}[ht!]
\begin{minipage}[t]{.485\linewidth}
    \centering
    \includegraphics[width=\linewidth,trim={0 0 0 0},clip]{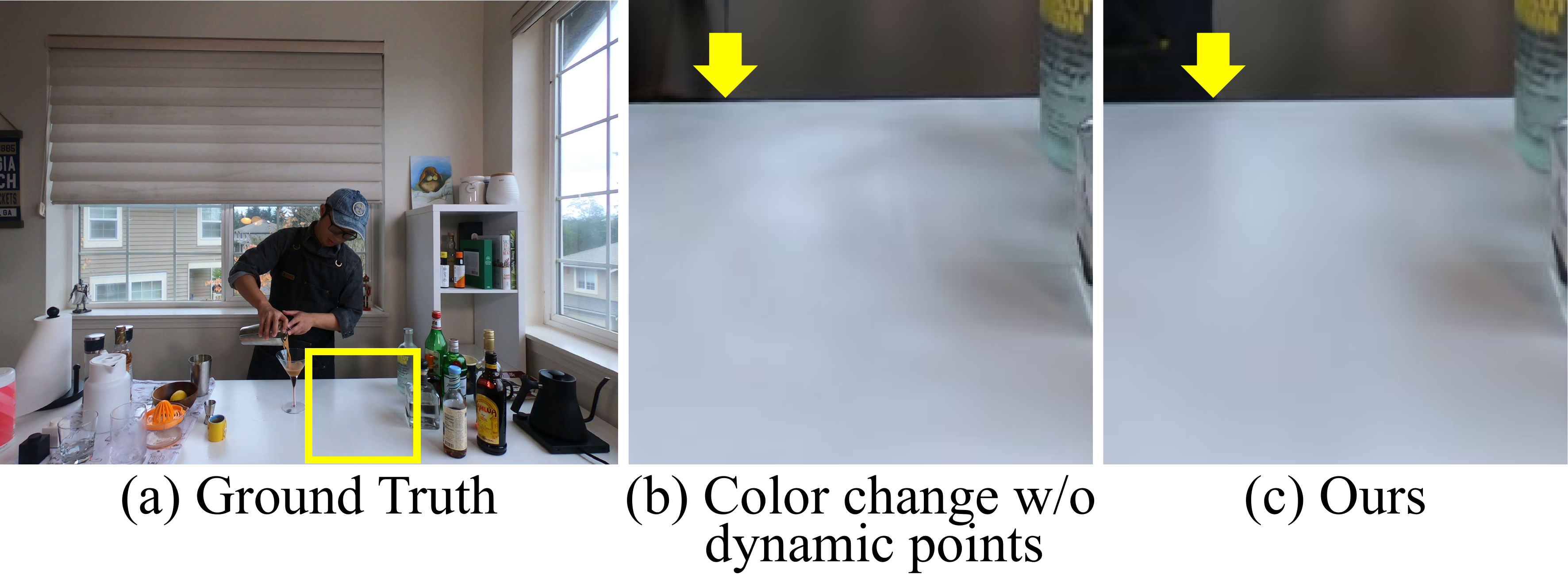}
    \vspace{-5mm}
    \caption{Comparison between (b) handling color changes without dynamic points and (c) our complete model.}
    \label{fig:appendix_withoutcolorchange}
\end{minipage}\hfill%
\begin{minipage}[t]{.485\linewidth}
    \centering
    \includegraphics[width=\linewidth,trim={0 0 0 0},clip]{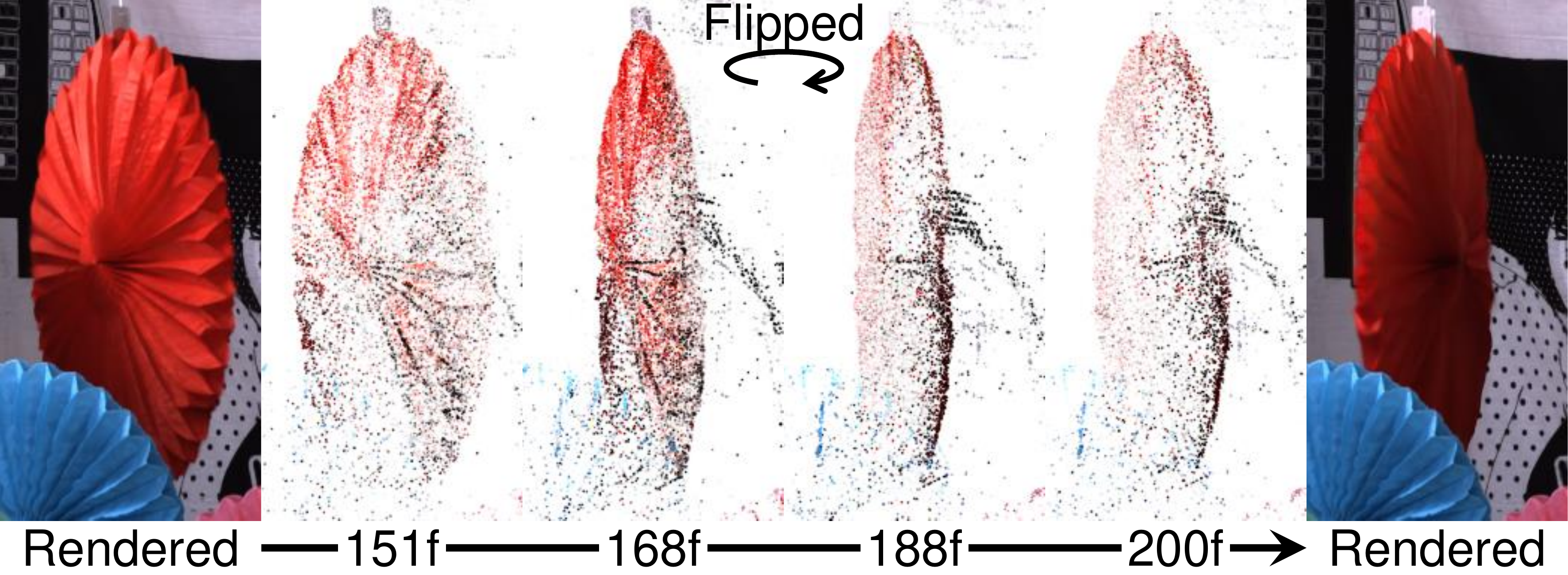}
    \vspace{-5mm}
    \caption{Visualization of the rotating decoration in the Technicolor Birthday scene.}
    \label{fig:appendix_rotatingdecoration}
\end{minipage} 
\vspace{-0mm}
\end{figure*}

\begin{wraptable}{r}{0.46\linewidth}
\vspace{-4mm}
\centering
\caption{Comparison results between without handling color changes and our complete model.}
\vspace{-1.5mm}
\resizebox{\linewidth}{!}{%
\begin{tabular}{cccc}
\toprule
~~~~~~~~Model~~~~~~~~ & ~~PSNR~~ & ~~SSIM$_1$\!~~ & ~~LPIPS~~ \\ \midrule 
3DGS~\cite{kerbl20233d}& \ycell 21.69 & \ycell 0.851 & \ycell 0.126 \\
3DGS + \tbf{Our dynamic} & \ocell 26.07 & \ocell 0.891 & \ocell 0.089 \\
\tbf{Ours}          & \rcell 28.79 & \rcell 0.912 & \rcell 0.070 \\
\bottomrule
\end{tabular}%
}
\label{tab:appendix_wo_color}
\end{wraptable}
We conduct experiments on the Coffee Martini scene from the Neural 3D Video dataset, focusing on scenarios where objects remain stationary but change color. We mask only the moving parts of the objects and train static Gaussians on the remaining regions using the original 3DGS model. The qualitative and quantitative results are presented in~\cref{fig:appendix_withoutcolorchange} and~\cref{tab:appendix_wo_color}. As shown in~\cref{fig:appendix_withoutcolorchange}, static points cannot handle changes such as shadows. However, Ex4DGS effectively manages these changes using dynamic points. In~\cref{tab:appendix_wo_color}, "3DGS" denotes the unmodified 3DGS results. "3DGS + Our dynamic" denotes the results when the dynamic regions are replaced with Ex4DGS. Even when only color changes occur without any movement, significant performance loss occurs if these changes are not treated as dynamic points.

\subsection{Reappearing Objects}
\label{appendix_repeative}
We visualize how Ex4DGS handles when objects disappear and reappear in~\cref{fig:appendix_rotatingdecoration}. \cref{fig:appendix_rotatingdecoration} illustrates the points associated with the decoration in the Birthday scene from the Technicolor dataset. It shows that the Gaussians corresponding to the part of the decoration that disappears at frame $\#168$ and reappears at frame $\#188$ have different distributions after the decoration flips that suggests that reappearing objects are regarded as new objects.

\subsection{Keyframe Interval Selections}
\label{appendix_keyframe_sel}

\begin{table}[ht!]
\caption{Ablation studies of keyframe interval selections and skipped frames.}
\vspace{1mm}
\centering
\resizebox{\linewidth}{!}{%
\begin{tabular}{c c cccc  c cccc c cccc c}
\toprule
Skipped frames & & \multicolumn{4}{c}{1} & & \multicolumn{4}{c}{2} & & \multicolumn{4}{c}{4} & \!\!\!\! \\ \cmidrule(lr){1-1} \cmidrule(){3-6} \cmidrule(){8-11} \cmidrule(){13-16}
~Keyframe interval~ & & \!\!PSNR\!\! & \!\!SSIM$_1$\!\!\! & \!\!LPIPS\!\! & \!\!\!Size(MB)\!\!\! & & \!\!PSNR\!\! & \!\!SSIM$_1$\!\!\! & \!\!LPIPS\!\! & \!\!\!Size(MB)\!\!\! & & \!\!PSNR\!\! & \!\!SSIM$_1$\!\!\! & \!\!LPIPS\!\! & \!\!\!Size(MB)\!\!\! \\ \midrule
1  & &        31.17 &        0.948 &        0.057 &        595 & &        31.47 &        0.948 &        0.056 &        415 & & \ycell 31.81 &        0.946 &        0.051 &        142   \\
2  & &        32.06 &        0.952 &        0.051 &        314 & & \ocell 32.33 & \ocell 0.954 & \ycell 0.049 &        322 & & \ycell 31.81 & \ocell 0.953 & \ocell 0.044 &        101   \\
5  & &        31.70 & \ycell 0.953 & \ycell 0.047 &        206 & & \ycell 32.29 & \ocell 0.954 & \rcell 0.043 &        126 & & \rcell 32.53 & \rcell 0.954 & \rcell 0.045 & \ycell 80    \\
10 & & \rcell 33.04 & \rcell 0.956 & \rcell 0.041 & \ycell 119 & & \rcell 32.65 & \rcell 0.956 & \rcell 0.043 & \ycell 93  & &        31.79 & \ocell 0.953 & \ycell 0.046 & \ocell 74  \\
20 & & \ocell 32.78 & \ocell 0.955 & \ocell 0.043 & \ocell 90  & &        32.07 & \ycell 0.952 & \ocell 0.047 & \ocell 78  & & \ocell 32.08 & \ocell 0.953 &        0.048 & \rcell 73  \\
50 & & \ycell 32.14 & \ocell 0.955 &        0.046 & \rcell 79  & &        31.93 &        0.951 &        0.052 & \rcell 72  & &        30.91 & \ycell 0.949 &        0.056 & \rcell 73 \\
\bottomrule
\end{tabular}%
}
\vspace{-1mm}
\label{tab:diff_keyframe}
\end{table}
In~\cref{tab:diff_keyframe}, we present results on the effects of keyframe intervals and motion magnitude on Cook Spinach scene from Neural 3D Video dataset. To simulate different motion speeds, we deliberately skip frames in the videos. As shown in~\cref{tab:diff_keyframe}, a keyframe interval of $10$ generally yields good results under most conditions. Smaller keyframe intervals tend to perform poorly, especially when the motion speed is low (i.e., fewer frames are skipped). However, as the skipped frame size increases smaller keyframe intervals begin to show better performance. It also shows that the model size decreases as the keyframe interval increases.

\subsection{Different Dynamic Point Conversion Rates}
\label{appendix_diff_conversion}

\begin{wraptable}{r}{0.46\linewidth}
\vspace{-4.5mm}
\centering
\caption{Ablation studies of dynamic point conversion rate.}
\vspace{-1.5mm}
\resizebox{\linewidth}{!}{%
\begin{tabular}{ccccc}
\toprule
Percentage & PSNR  & SSIM$_1$\! & LPIPS & \!\!\!Size(MB)\!\!\! \\ \midrule 
0.5     &        32.36 & \ocell 0.955 & \ocell 0.043 & \rcell 103       \\
1       & \ycell 32.48 & \rcell 0.956 & \ycell 0.044 & \ocell 115       \\
2       & \rcell 33.04 & \rcell 0.956 & \rcell 0.041 & \ycell 119       \\
4       & \ocell 32.89 & \ocell 0.955 &        0.045 &        227       \\
8       &        31.33 & \ycell 0.954 &        0.048 &        367 \\
\bottomrule
\end{tabular}%
}
\label{tab:diff_conversion}
\vspace{-5mm}
\end{wraptable}
We experiment with different dynamic point conversion rates on Cook Spinach scene from Neural 3D Video dataset in~\cref{tab:diff_conversion}. Our results indicate that the best performance is achieved when the extraction percentage is set to $2\%$. If the percentage is too low, not enough dynamic points will be extracted; conversely, if it is too high, too many dynamic points may be extracted, leading to overfitting and degraded performance.

\section{Additional Comparisons}
\label{appendix_more_comparison}
To assess the robustness of Ex4DGS, we sample different frame intervals from the Technicolor dataset. First, we experiment with occlusion scenarios in~\cref{appendix_handling_occlusion}. Next, we present results when a new object appears in~\cref{appendix_handling_newly}. Finally, we train on an extremely long-duration video in~\cref{appendix_longer_duration}.

\subsection{Handling Occlusion}
\label{appendix_handling_occlusion}

\begin{figure*}[h]
\vspace{-1mm}
\centering
\includegraphics[width=\linewidth,trim={0 0 0 0},clip]{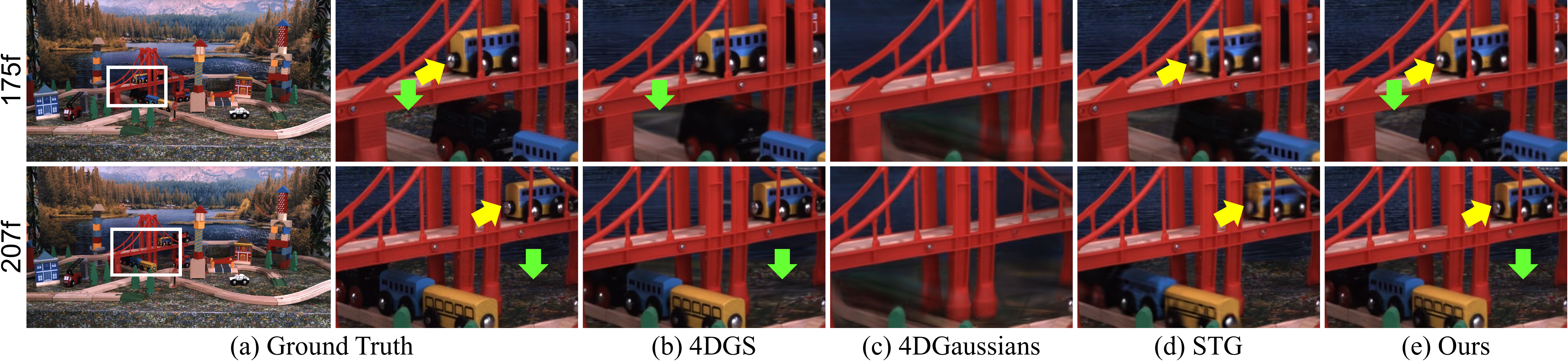}
\vspace{-5mm}
\caption{Qualitative comparison of the repeatedly occluded objects in the Technicolor Train scene over a sequence of 100 frames (frame \#170 to \#269). All models are trained with the point cloud data from the frame \#170.}
\label{fig:appendix_occuluded}
\end{figure*}

\begin{wraptable}{r}{0.46\linewidth}
\vspace{-4.5mm}
\centering
\caption{\textls[-8]{Quantitative results of the repeatedly occluded objects in the Technicolor Train scene.}}
\vspace{-1.5mm}
\resizebox{\linewidth}{!}{%
\begin{tabular}{cccc}
\toprule
~~~~~~~~Model~~~~~~~~ & ~~PSNR~~ & ~~SSIM$_1$\!~~ & ~~LPIPS~~ \\ \midrule 
STG~\cite{li2023spacetime}& \ocell 32.17 & \ocell 0.940 & \rcell 0.035 \\
4DGS~\cite{yang2023real}& \ycell 29.11 & \ycell 0.877 & \ycell 0.119 \\
4DGaussians~\cite{wu20234dgaussians}& 23.31        & 0.657        & 0.385        \\
\tbf{Ours}             & \rcell 32.24 & \rcell 0.941 & \ocell 0.044 \\
\bottomrule
\end{tabular}%
}
\label{tab:appendix_occuluded}
\vspace{-2mm}
\end{wraptable}
We sample 100 frames (frames \#170 to \#269) from the Train scene in the Technicolor dataset containing occlusions of dynamic objects and compare Ex4DGS with other models. We use the point cloud prior of the first frame, which provides no information about the reappearing object after the occlusion. We compare the performance of STG, 4DGS and 4D Gaussians models in~\cref{tab:appendix_occuluded} and~\cref{fig:appendix_occuluded}. In these results, the explicit-based models, STG, 4DGS, and ours, perform significantly better. In the case of STG, the dynamic part is not well learned as the frames progress, and while 4DGS can render the dynamic part effectively, it struggles with the static part, negatively affecting the overall performance. In particular, 4D Gaussians, being an implicit model, fails to disentangle the static and dynamic components, resulting in missing renderings of the dynamic part. Our model, on the other hand, performs well and effectively learns both static and dynamic parts.

\subsection{Handling Newly Appearing Objects}
\label{appendix_handling_newly}

\begin{figure*}[t]
    \centering
    \includegraphics[width=\linewidth]{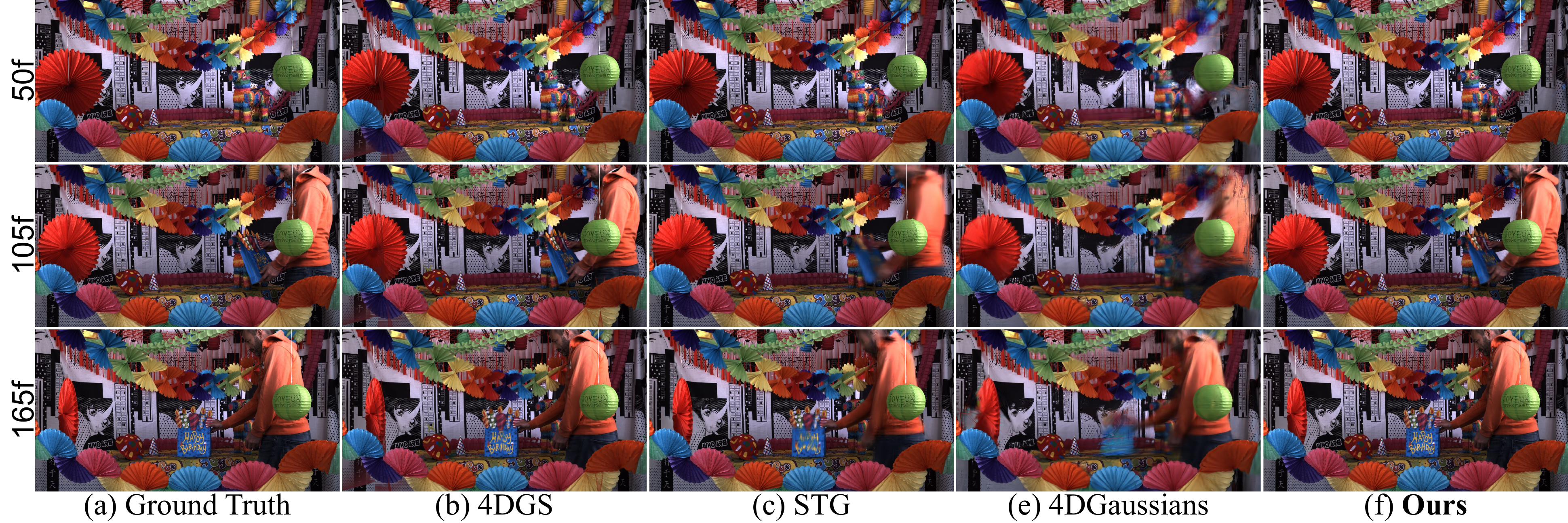}
    \vspace{-5mm}
    \caption{Qualitative comparison of the appearing objects in Technicolor Birthday scene over a sequence of 120 frames (frame \#50 to \#169). All models are trained with the point cloud data from the frame \#50.
    }
    \label{fig:appendix_newobject}
    \vspace{-1mm}
\end{figure*}

\begin{wraptable}{r}{0.46\linewidth}
\vspace{-4.5mm}
\centering
\caption{Quantitative results of the appearing objects in Technicolor Birthday scene.}
\vspace{-1.5mm}
\resizebox{\linewidth}{!}{%
\begin{tabular}{cccc}
\toprule
~~~~~~~~Model~~~~~~~~ & ~~PSNR~~ & ~~SSIM$_1$\!~~ & ~~LPIPS~~ \\ \midrule 
STG~\cite{li2023spacetime}& \ycell 27.62 & \ycell 0.903 & \ocell 0.080 \\
4DGS~\cite{yang2023real}& \ocell 28.69 & \ocell 0.907 & \ycell 0.086 \\
4DGaussians~\cite{wu20234dgaussians}&        21.51 &        0.712 &        0.291 \\
\tbf{Ours}   & \rcell 30.56 & \rcell 0.929 & \rcell 0.051 \\
\bottomrule
\end{tabular}%
}
\vspace{-3mm}
\label{tab:appendix_newly}
\end{wraptable}
We conduct an experiment to determine whether Ex4DGS can learn about newly appearing objects that require the splitting of dynamic components. We sample 120 frames (frame \#50 to \#169) from the Birthday scene in the Technicolor dataset, during which a person appears. All models use a point cloud prior from a frame where the person is not yet visible. The numerical results are presented in~\cref{tab:appendix_newly}, and the rendered images are shown in~\cref{fig:appendix_newobject}. In contrast With the assertion made in the conclusion, the result is indeed feasible because Gaussians from neighboring objects can be utilized to facilitate the splitting process, even in the case of newly appearing objects. This is due to the effectiveness of the proposed splitting pipeline for static and dynamic Gaussians, which can handle newly appearing objects even when no initial Gaussian is provided.

\subsection{Extremely Long Duration}
\label{appendix_longer_duration}

\begin{figure*}[h!]
\vspace{-5mm}
    \captionsetup[subfloat]{captionskip=2pt}
    \centering
    \subfloat[Animated video]{\animategraphics[width=0.195\linewidth,loop,autoplay]{10}{figures/long_duration_gif/}{000}{200}%
    \label{fig:ablation_longduration_a}}
    \hfil
    \subfloat[1st frame]{\includegraphics[width=0.195\linewidth,trim={0 0 0 0},clip]{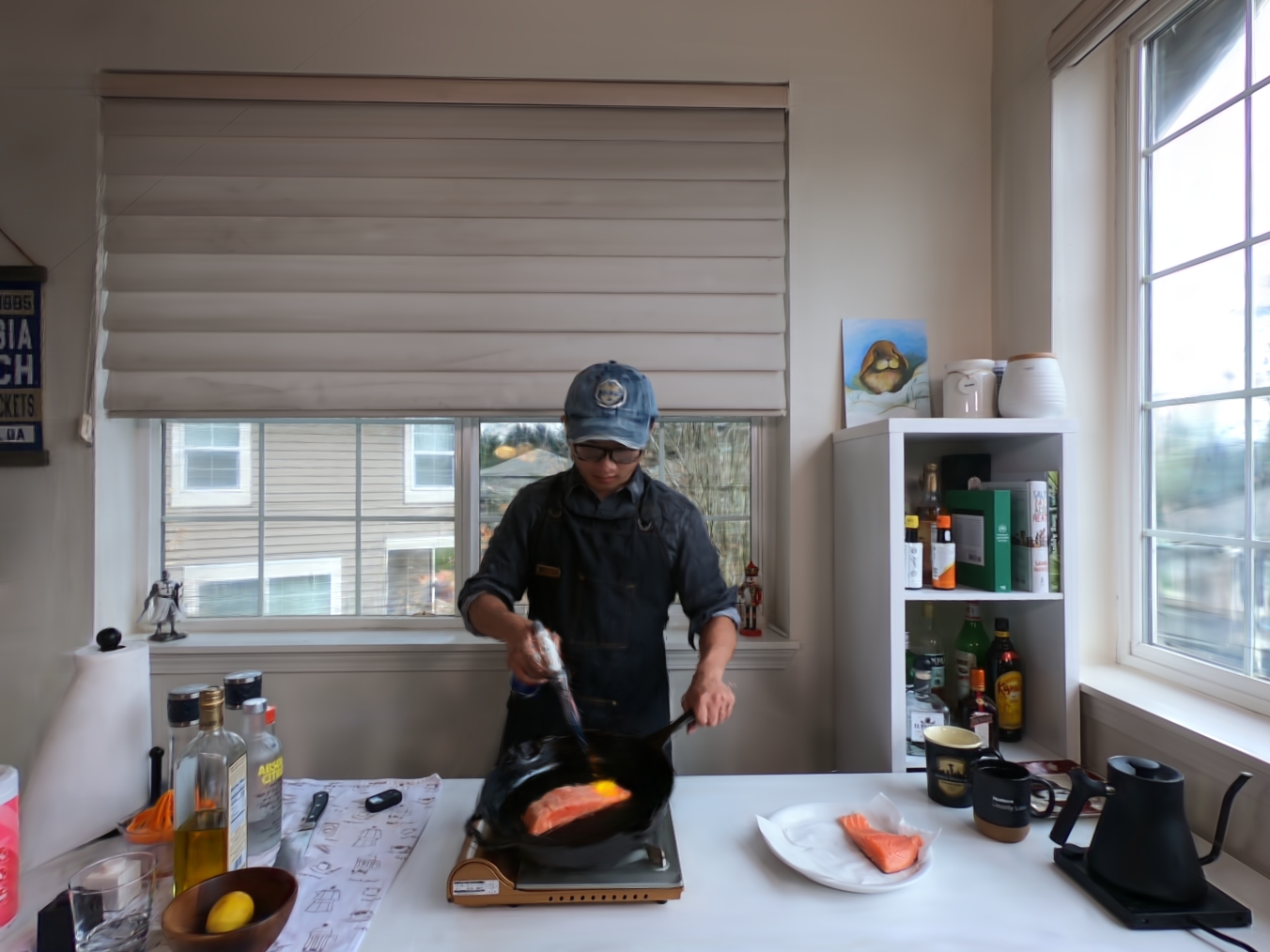}%
    \label{fig:ablation_longduration_b}}
    \hfil
    \subfloat[167th frame]{\includegraphics[width=0.195\linewidth,trim={0 0 0 0},clip]{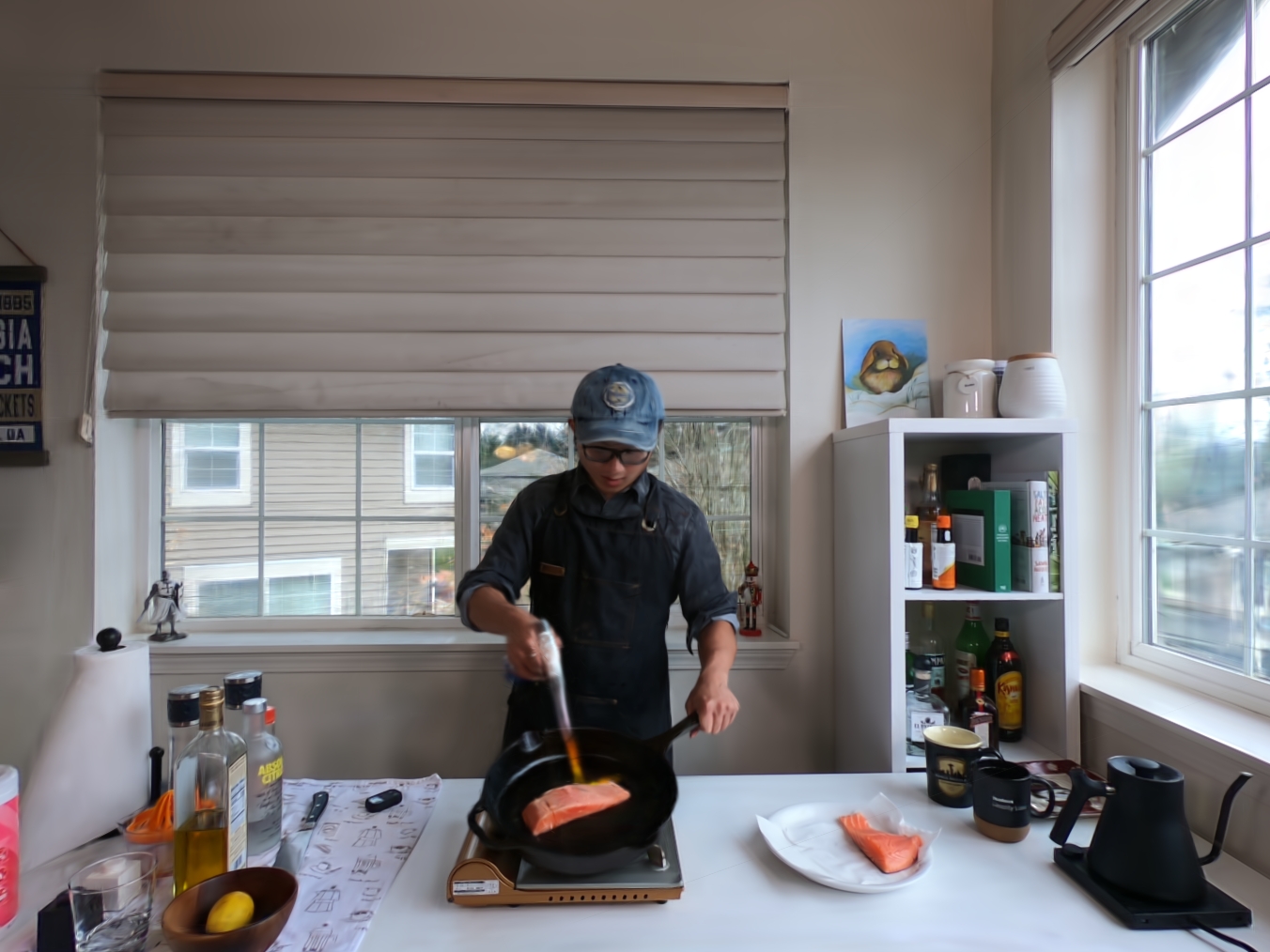}%
    \label{fig:ablation_longduration_c}}
    \hfil
    \subfloat[333th frame]{\includegraphics[width=0.195\linewidth,trim={0 0 0 0},clip]{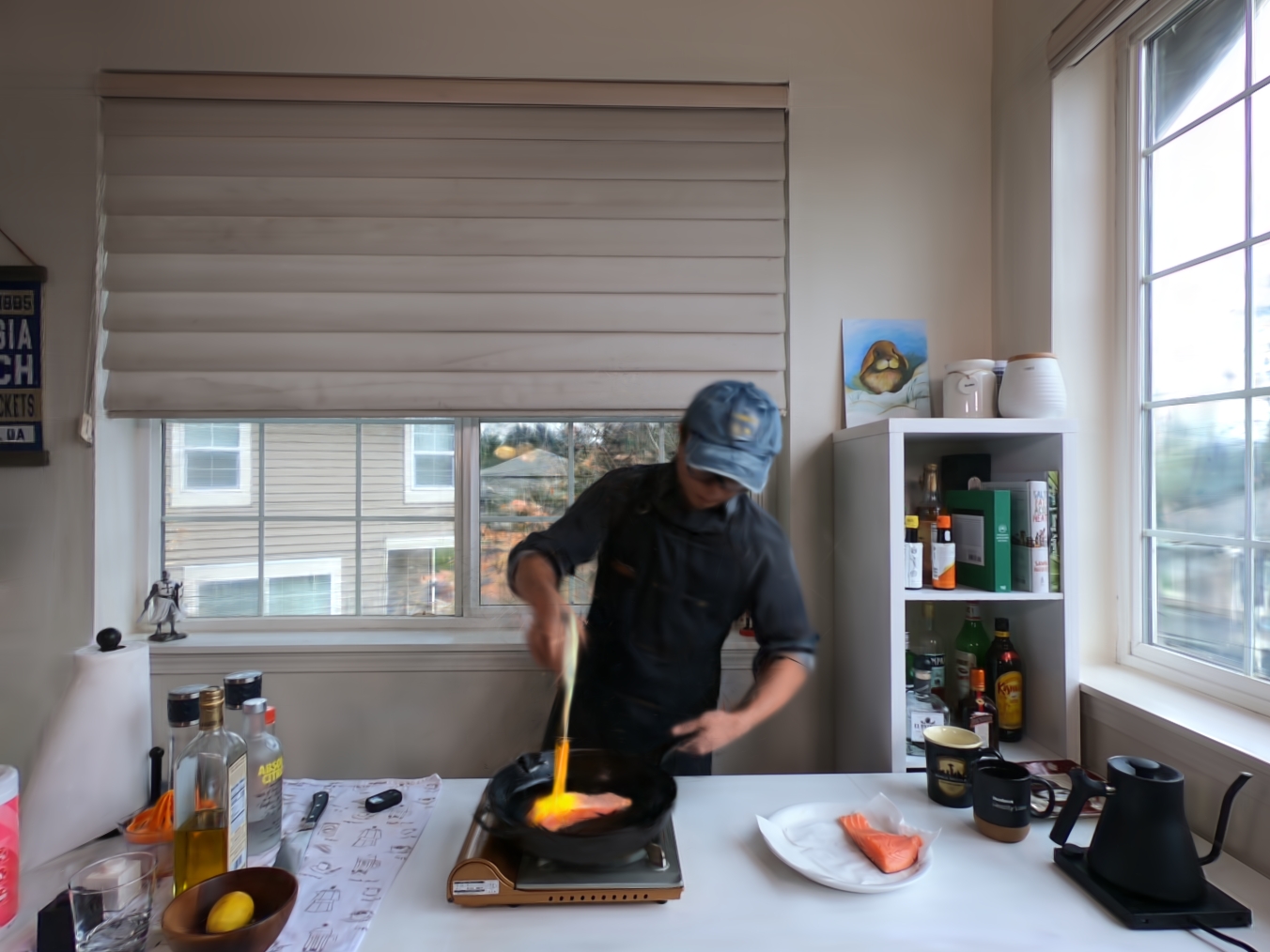}%
    \label{fig:ablation_longduration_d}}
    \hfil
    \subfloat[500th frame]{\includegraphics[width=0.195\linewidth,trim={0 0 0 0},clip]{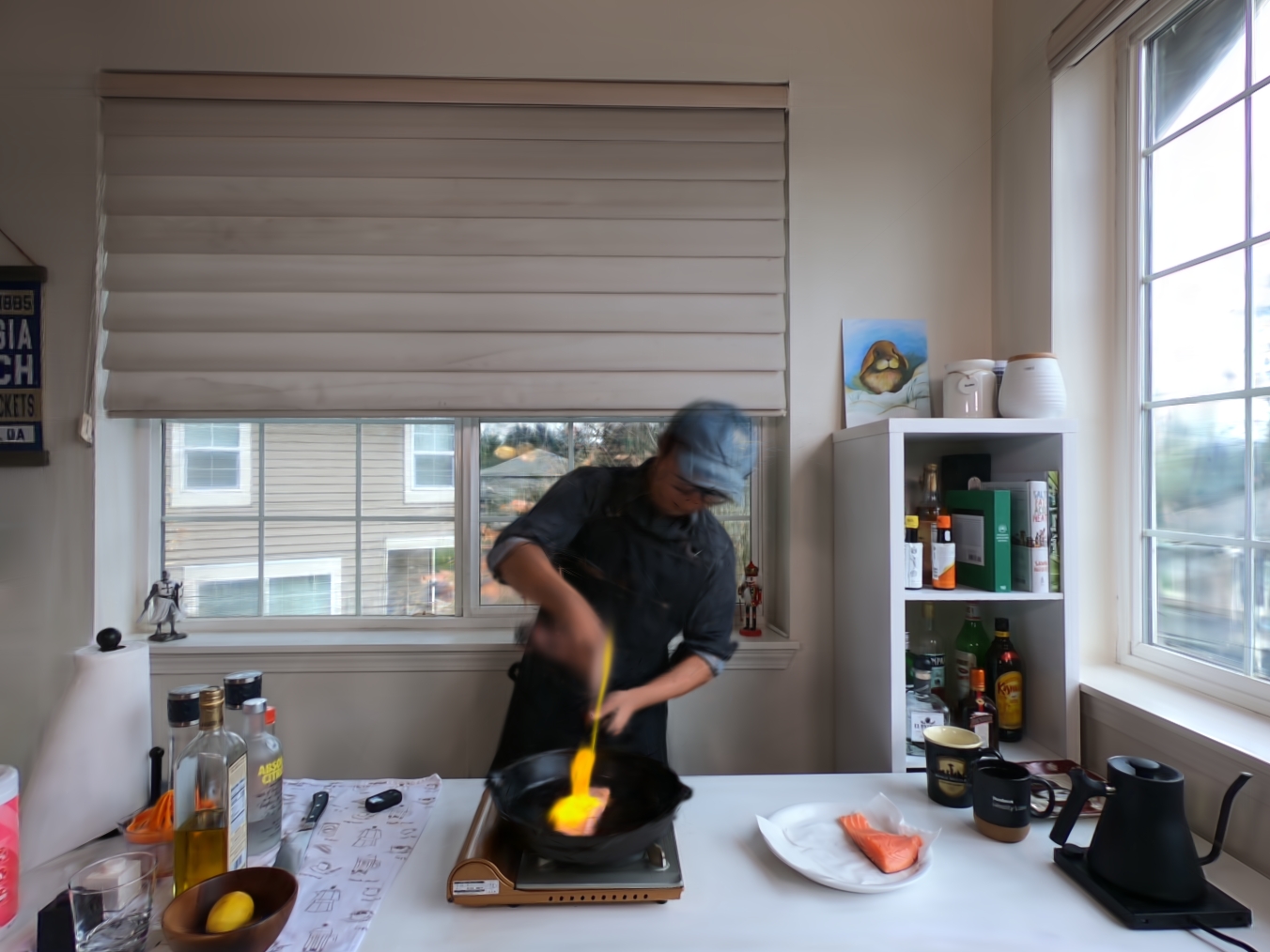}%
    \label{fig:ablation_longduration_e}}\\\vspace{-3mm}
    \subfloat[667th frame]{\includegraphics[width=0.195\linewidth,trim={0 0 0 0},clip]{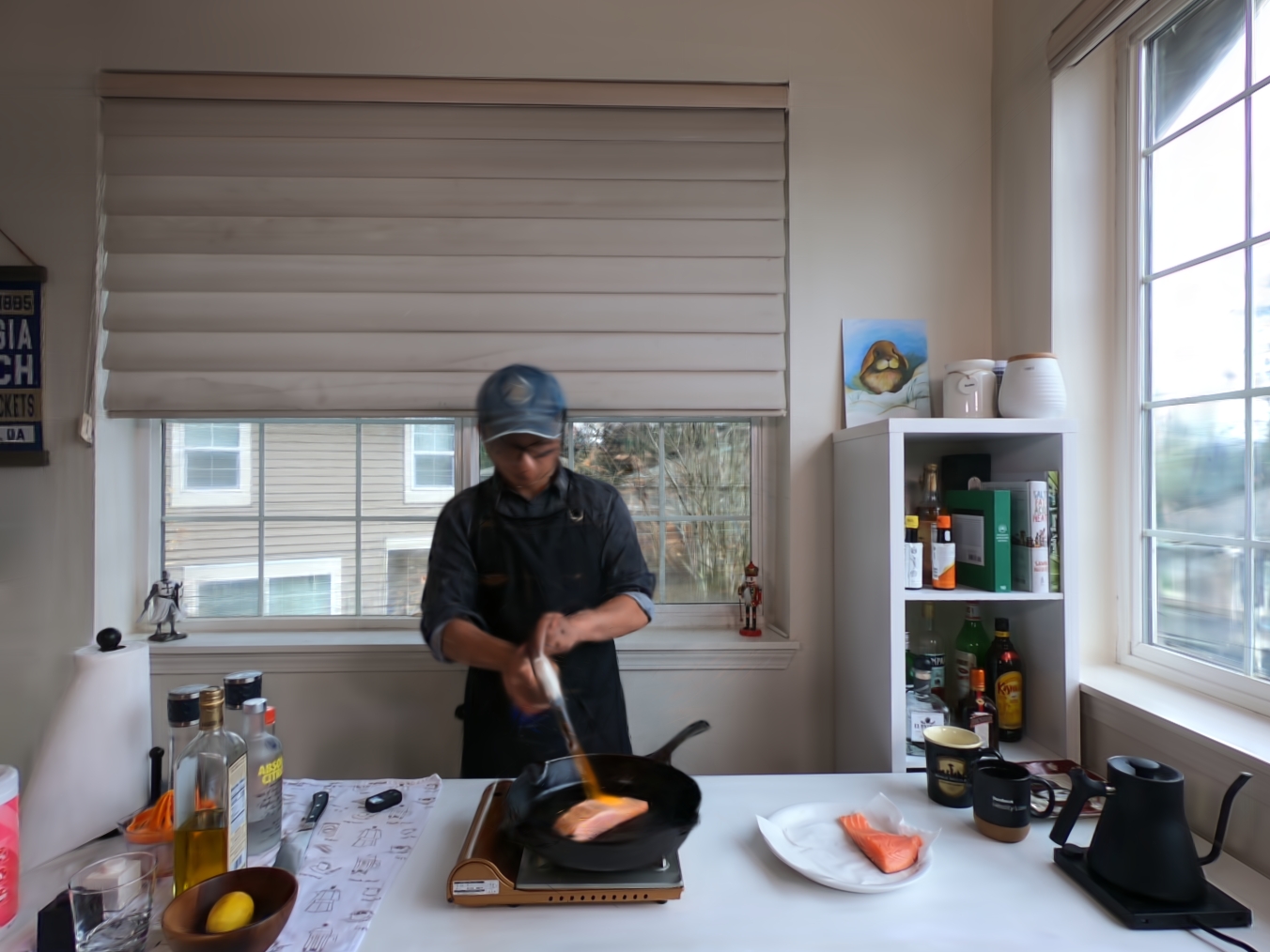}%
    \label{fig:ablation_longduration_f}}
    \hfil
    \subfloat[833th frame]{\includegraphics[width=0.195\linewidth,trim={0 0 0 0},clip]{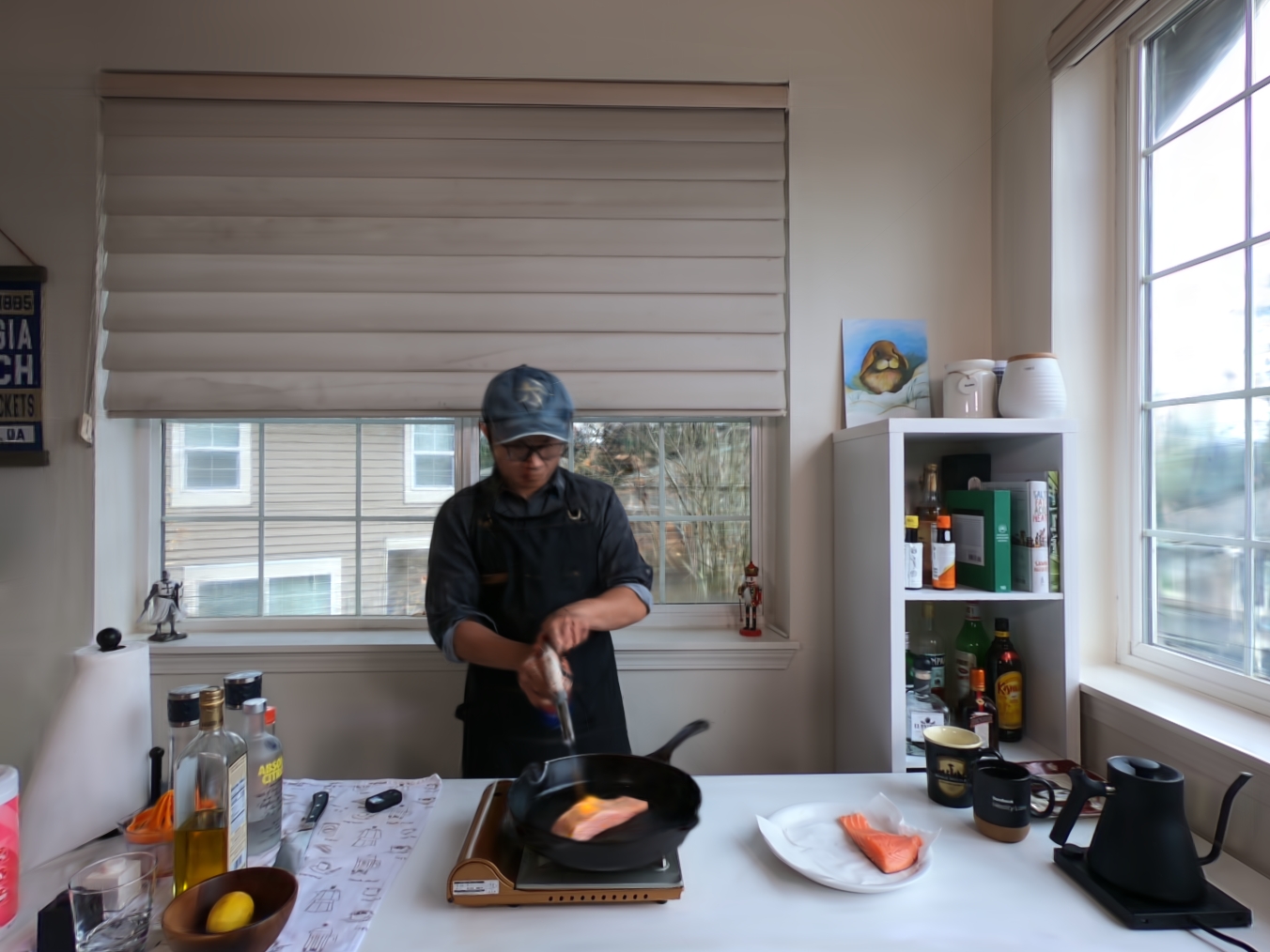}%
    \label{fig:ablation_longduration_g}}
    \hfil
    \subfloat[1000th frame]{\includegraphics[width=0.195\linewidth,trim={0 0 0 0},clip]{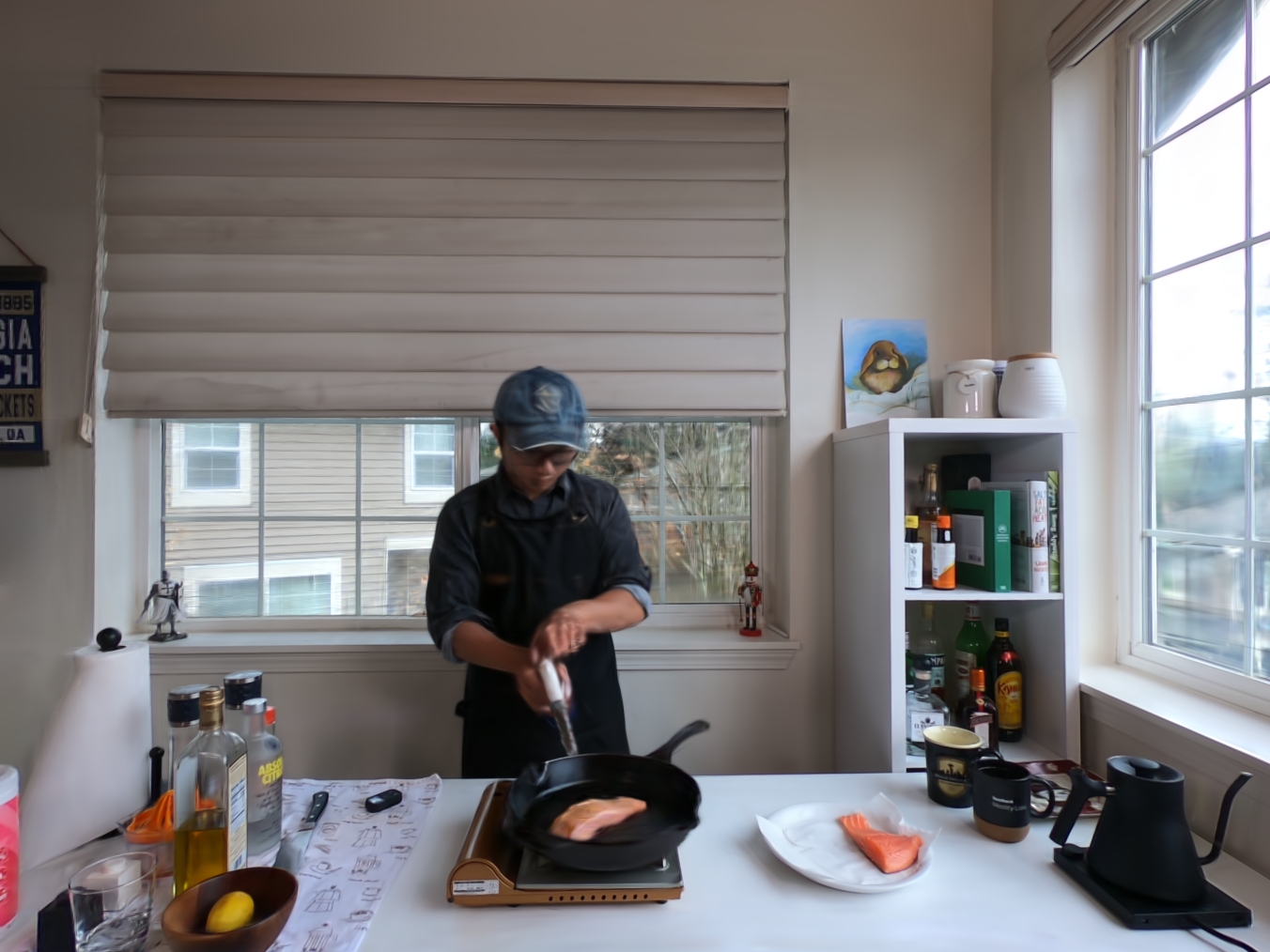}%
    \label{fig:ablation_longduration_h}}
    \hfil
    \subfloat[1000th static]{\includegraphics[width=0.195\linewidth,trim={0 0 0 0},clip]{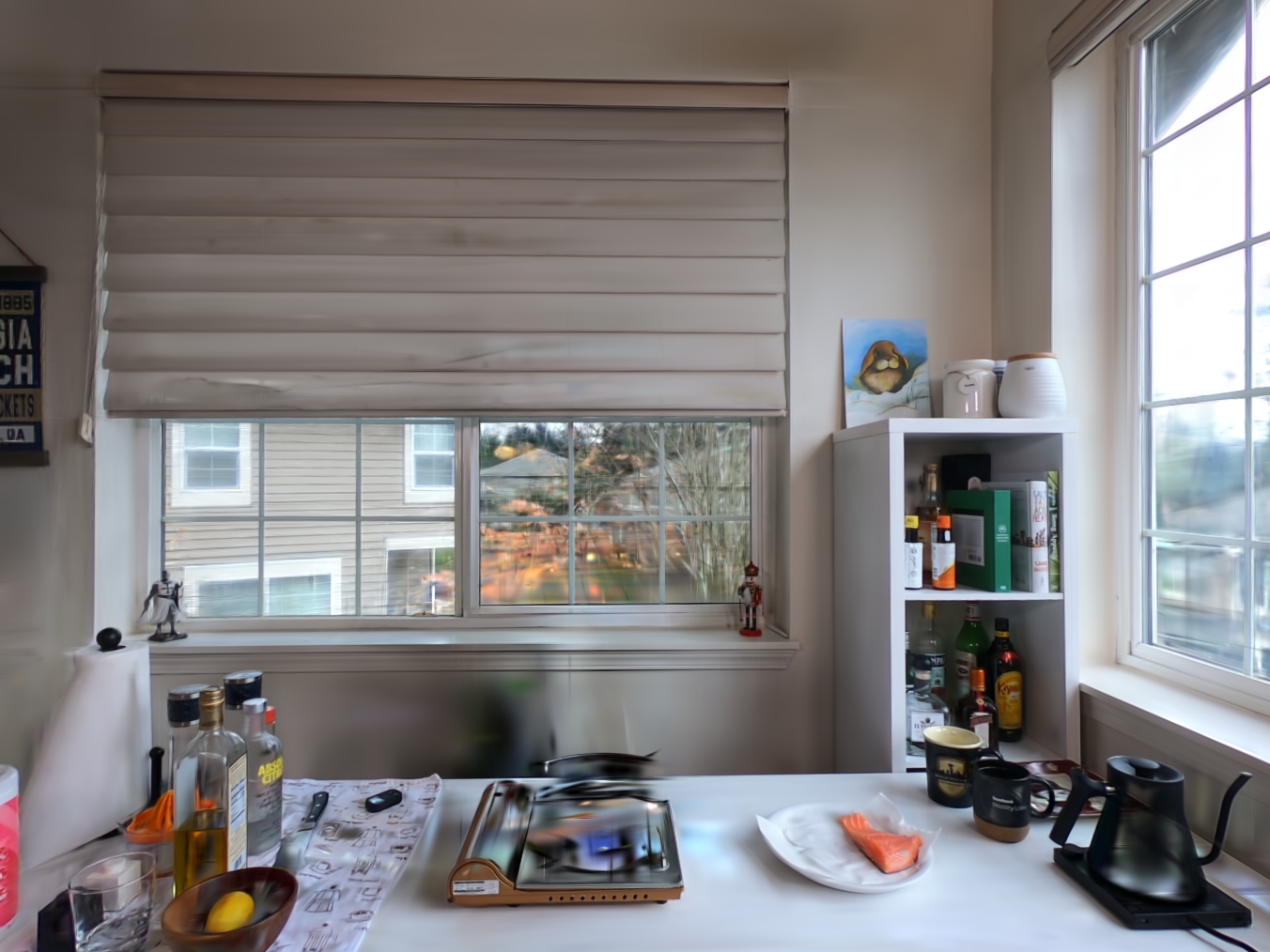}%
    \label{fig:ablation_longduration_i}}
    \hfil
    \subfloat[1000th dynamic]{\includegraphics[width=0.195\linewidth,trim={0 0 0 0},clip]{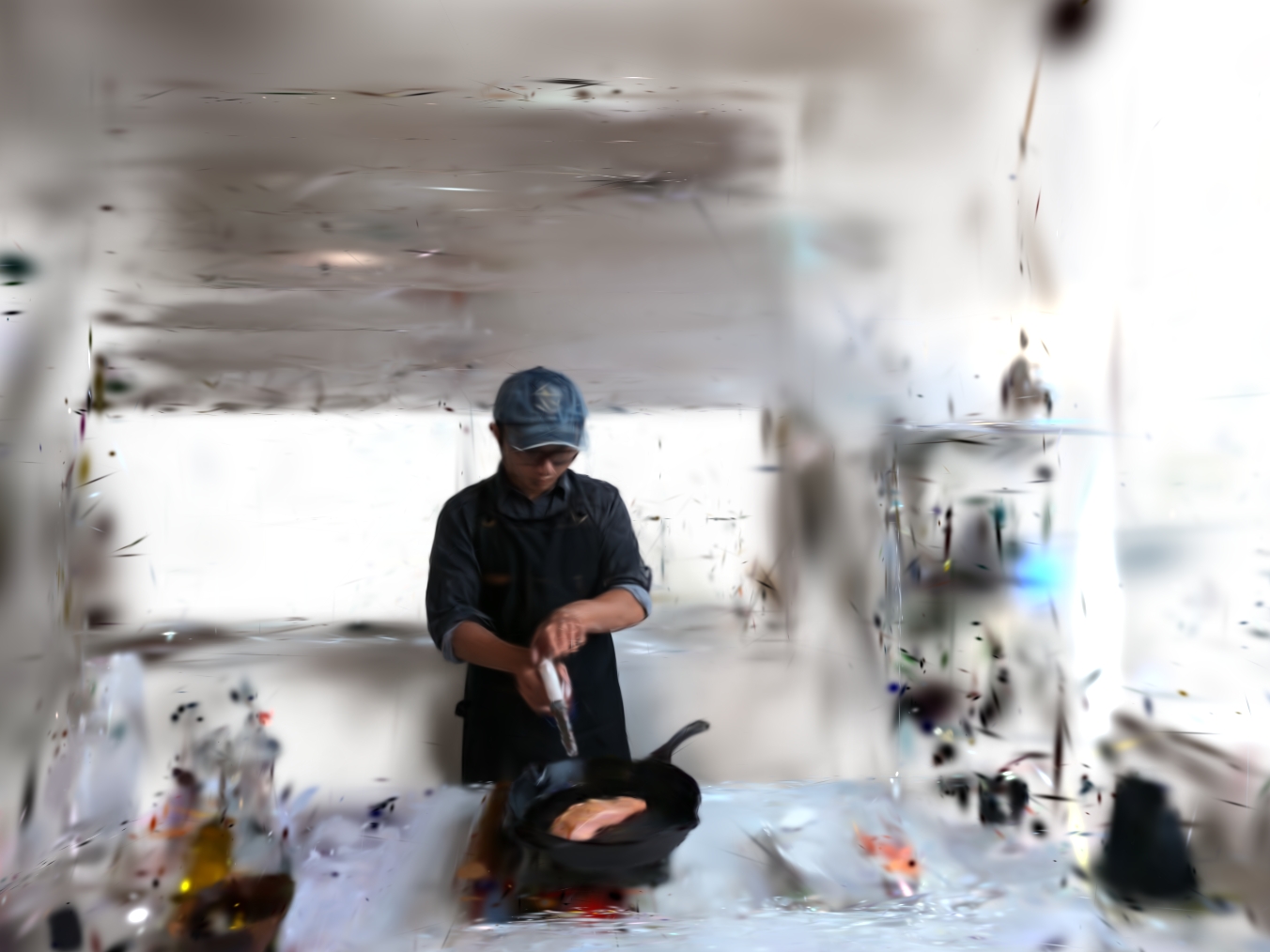}%
    \label{fig:ablation_longduration_j}}
    \vspace{-1mm}
    \caption{Evaluation of the extremely long video on Flame Salmon scene in Neural 3D Video dataset. Best viewed at \textit{Adobe\,Acrobat\,Reader}.}
    \label{fig:appendix_long_flamesalmon}
\end{figure*}

\begin{wraptable}{r}{0.46\linewidth}
\vspace{-4.5mm}
\centering
\caption{Quantitative results of the extremely long video on Flame Salmon scene in Neural 3D Video dataset.}
\vspace{-1.5mm}
\resizebox{\linewidth}{!}{%
\begin{tabular}{ccccc}
\toprule
Model & PSNR & SSIM$_1$\! & LPIPS & \!\!\!Size(MB)\!\!\!\\ \midrule 
4DGS~\cite{yang2023real}& \ycell 26.26 & \ycell 0.897 & \ycell 0.115 & \ycell 6331 \\
4DGaussians~\cite{wu20234dgaussians}& \ocell 28.37 & \ocell 0.903 & \ocell 0.097 & \rcell 75  \\
\tbf{Ours}      & \rcell 28.77 & \rcell 0.919 & \rcell 0.076 & \ocell 392 \\
\bottomrule
\end{tabular}%
}
\label{tab:appendix_long_flamesalmon}
\vspace{-4mm}
\end{wraptable}

We conduct an experiment using a longer sequence of frames (1,000 frames, 20,000 images in total) on the Flame Salmon scene from the Neural 3D Video dataset. The results are presented in~\cref{tab:appendix_long_flamesalmon}, and the rendered images are shown in~\cref{fig:appendix_long_flamesalmon}. The results of this experiment demonstrate that our model is capable of effective learning with reasonable storage requirements, even for extremely long videos. While the 4D Gaussian model produces acceptable results, its performance declines in areas where new objects, such as flames, appear. This indicates that the rendering quality may vary depending on the presence or absence of newly appearing objects, as discussed in~\cref{appendix_handling_occlusion} and~\cref{appendix_handling_newly}.

\vspace{-1mm}
\section{Detailed Results}
\label{appendix_deatil_results}
\vspace{-1mm}

In this section, we report the scene breakdown results of PSNR, SSIM$_1$, SSIM$_2$, and LPIPS on the Technicolor dataset and SSIM$_1$, SSIM$_2$ and LPIPS on the Neural 3D Video dataset.

\begin{table}[h]
\vspace{5mm}
\caption{Per-scene quantitative comparison on Technicolor dataset. {\small\textdagger}: Trained with sparse point cloud input.}
\vspace{3mm}
\centering
\resizebox{0.85\linewidth}{!}{%
\begin{tabular}{ccccccc}
\toprule
\multirow{2}{*}{Model\vspace{-5pt}} & \multicolumn{6}{c}{PSNR} \\ \cmidrule(lr){2-7} 
 & Birthday & \nsp\,~Fabien~\,\nsp & \psp~Painter~\psp & \,Theater\, & \,~~Train~~\, & \psp Average\psp \\ \midrule
DyNeRF~\cite{li2022neural}&        29.20 &        32.76 & \ycell 35.95 &        29.53 & \ycell 31.58 &        31.80 \\
HyperReel~\cite{hyperreel}&        29.99 & \ocell 34.70 &        35.91 & \rcell 33.32 &        29.74 & \ycell 32.73 \\
STG\textsuperscript{\textdagger}~\cite{li2023spacetime}& \ocell 31.96 & \ycell 34.53 & \ocell 36.47 &        30.54 & \rcell 32.65 & \ocell 33.23 \\
4DGS~\cite{yang2023real}&        28.01 &        26.19 &        33.91 & \ycell 31.62 &        27.96 &        29.54 \\
~~4D Gaussians~\cite{wu20234dgaussians}~~& \ycell 30.87 &        33.56 &        34.36 &        29.81 &        25.35 &        30.79 \\
\tbf{Ours}   & \rcell 32.38 & \rcell 35.38 & \rcell 36.73 & \ocell 31.84 & \ocell 31.77 & \rcell 33.62 \\
\bottomrule
\label{tab:Techni_PSNR}
\end{tabular}%
}

\resizebox{0.85\linewidth}{!}{%
\begin{tabular}{ccccccc}
\toprule
\multirow{2}{*}{Model\vspace{-5pt}} & \multicolumn{6}{c}{SSIM$_1$} \\ \cmidrule(lr){2-7} 
 & Birthday & \nsp\,~Fabien~\,\nsp & \psp~Painter~\psp & \,Theater\, & \,~~Train~~\, & \psp Average\psp \\ \midrule
HyperReel~\cite{hyperreel}& \ycell 0.922 & \rcell 0.895 & \ocell 0.923 & \rcell 0.895 & \ycell 0.895 & \ycell 0.906 \\
STG\textsuperscript{\textdagger}~\cite{li2023spacetime}& \ocell 0.942 & \ycell 0.877 & \ocell 0.923 & \ycell 0.872 & \rcell 0.945 & \ocell 0.912 \\
4DGS~\cite{yang2023real}&        0.902 &        0.856 & \ycell 0.897 &        0.869 &        0.843 &        0.873 \\
~~4D Gaussians~\cite{wu20234dgaussians}~~&       0.904 &        0.854 &        0.884 &        0.841 &        0.730 &        0.843 \\
\tbf{Ours}   & \rcell 0.943 & \ocell 0.889 & \rcell 0.929 & \ocell 0.880 & \ocell 0.937 & \rcell 0.916 \\
\bottomrule
\label{tab:Techni_SSIM1}
\end{tabular}%
}

\resizebox{0.85\linewidth}{!}{%
\begin{tabular}{ccccccc}
\toprule
\multirow{2}{*}{Model\vspace{-5pt}} & \multicolumn{6}{c}{SSIM$_2$} \\ \cmidrule(lr){2-7} 
 & Birthday & \nsp\,~Fabien~\,\nsp & \psp~Painter~\psp & \,Theater\, & \,~~Train~~\, & \psp Average\psp \\ \midrule
DyNeRF~\cite{li2022neural}& \ycell 0.952 & \rcell 0.965 & \rcell 0.972 & \ycell 0.939 & \ocell 0.962 & \ycell 0.958 \\
STG\textsuperscript{\textdagger}~\cite{li2023spacetime}& \ocell 0.969 & \ycell 0.955 & \ocell 0.970 & \ycell 0.939 & \rcell 0.967 & \ocell 0.960 \\
4DGS~\cite{yang2023real}&        0.944 &        0.943 & \ycell 0.957 & \ocell 0.940 &        0.901 &        0.937 \\
~~4D Gaussians~\cite{wu20234dgaussians}~~&        0.950 &        0.946 &        0.951 &        0.925 &        0.832 &        0.921 \\
\tbf{Ours}   & \rcell 0.970 & \ocell 0.961 & \rcell 0.972 & \rcell 0.944 & \ycell 0.961 & \rcell 0.962 \\
\bottomrule
\label{tab:Techni_SSIM2}
\end{tabular}%
}

\resizebox{0.85\linewidth}{!}{%
\begin{tabular}{ccccccc}
\toprule
\multirow{2}{*}{Model\vspace{-5pt}} & \multicolumn{6}{c}{LPIPS} \\ \cmidrule(lr){2-7} 
 & Birthday & \nsp\,~Fabien~\,\nsp & \psp~Painter~\psp & \,Theater\, & \,~~Train~~\, & \psp Average\psp \\ \midrule
DyNeRF~\cite{li2022neural}&        0.067 &        0.242 &        0.146 &        0.188 &        0.067 &        0.142 \\
HyperReel~\cite{hyperreel}& \ycell 0.053 & \ycell 0.186 & \ycell 0.117 & \rcell 0.115 & \ycell 0.072 & \ycell 0.109 \\
STG\textsuperscript{\textdagger}~\cite{li2023spacetime}& \rcell 0.039 & \ocell 0.134 & \ocell 0.097 & \ocell 0.121 & \rcell 0.033 & \rcell 0.085 \\
4DGS~\cite{yang2023real}&        0.089 &        0.197 &        0.136 &        0.155 &        0.166 &        0.149 \\
~~4D Gaussians~\cite{wu20234dgaussians}~~&        0.087 & \ycell 0.186 &        0.161 &        0.187 &        0.271 &        0.178 \\
\tbf{Ours}   & \ocell 0.044 & \rcell 0.123 & \rcell 0.091 & \ycell 0.129 & \ocell 0.052 & \ocell 0.088 \\
\bottomrule
\label{tab:Techni_LPIPS}
\end{tabular}%
}
\end{table}

\begin{table}[h]
\caption{Per-scene quantitative comparison on Neural 3D Video dataset. {\small\textdaggerdbl}: Trained using a dataset split into 150 frames.}
\centering
\vspace{1mm}
\resizebox{0.85\linewidth}{!}{%
\begin{tabular}{cccccccc}
\toprule
\multirow{2}{*}{Model\vspace{-15pt}} & \multicolumn{7}{c}{SSIM$_1$} \\ \cmidrule(lr){2-8} 
 & \tworow{Coffee}{Martini} & \tworow{Cook}{Spinach} & \tworow{\!\!\!\!\!Cut\,Roasted\!\!\!\!\!}{Beef} & \tworow{Flame}{Salmon} & \tworow{Flame}{~Steak~} & \tworow{Sear}{~Steak~} & Average \\ \midrule
NeRFPlayer~\cite{song2023nerfplayer}& 0.951         & 0.929        & 0.908         & 0.940        & 0.950       & 0.908      & 0.931   \\
HyperReel~\cite{hyperreel}& 0.892         & 0.941        & 0.945         & 0.882        & 0.949       & 0.952      & 0.927   \\ \midrule
\multicolumn{8}{c}{Dense COLMAP point cloud input}      \\
STG\textsuperscript{\textdaggerdbl}~\cite{li2023spacetime}& 0.916         & 0.952        & 0.954         & 0.918        & 0.960       & 0.961      & 0.944   \\
4DGS~\cite{yang2023real}& \NA           & \NA          & 0.980         & 0.960        & \NA         & \NA        & 0.970   \\
4DGaussians~\cite{wu20234dgaussians}& 0.905         & 0.949        & 0.957         & 0.917        & 0.954       & 0.957      & 0.940   \\ \midrule
\multicolumn{8}{c}{Sparse COLMAP point cloud input}      \\
STG\textsuperscript{\textdaggerdbl}~\cite{li2023spacetime}& \ocell 0.904  & \ycell 0.946 & \ycell 0.946  & \ocell 0.913 & \ocell 0.954 & \ocell 0.955 & \ocell 0.936 \\
4DGS~\cite{yang2023real}& \ycell 0.902  & \rcell 0.948 & \ocell 0.947  & \ycell 0.904 & \ocell 0.954 & \ocell 0.955 & \ycell 0.935 \\
4DGaussians~\cite{wu20234dgaussians}& 0.893         & 0.944        & 0.913         & 0.896        & \ycell 0.946 & \ycell 0.946 & 0.923   \\
\tbf{Ours}             & \rcell 0.915  & \ocell 0.947 & \rcell 0.948  & \rcell 0.917 & \rcell 0.956 & \rcell 0.959 & \rcell 0.940 \\
\bottomrule
\label{tab:N3V_SSIM1}
\end{tabular}%
}

\vspace{-1mm}

\resizebox{0.85\linewidth}{!}{%
\begin{tabular}{cccccccc}
\toprule
\multirow{2}{*}{Model\vspace{-15pt}} & \multicolumn{7}{c}{SSIM$_2$} \\ \cmidrule(lr){2-8} 
 & \tworow{Coffee}{Martini} & \tworow{Cook}{Spinach} & \tworow{\!\!\!\!\!Cut\,Roasted\!\!\!\!\!}{Beef} & \tworow{Flame}{Salmon} & \tworow{Flame}{~Steak~} & \tworow{Sear}{~Steak~} & Average \\ \midrule
Neural Volumes~\cite{lombardi2019neural}& \NA           & \NA          & \NA              & 0.876        & \NA         & \NA        & 0.876   \\
LLFF~\cite{mildenhall2019llff}& \NA           & \NA          & \NA              & 0.848        & \NA         & \NA        & 0.848   \\
DyNeRF~\cite{li2022neural}& \NA           & \NA          & \NA              & 0.960        & \NA         & \NA        & 0.960   \\
HexPlane~\cite{cao2023hexplane}& \NA           & 0.970        & 0.974            & 0.960        & 0.978       & 0.978      & 0.972   \\
K-Planes~\cite{fridovich2023k}& 0.953         & 0.966        & 0.966            & 0.953        & 0.970       & 0.974      & 0.964   \\
MixVoxels-L~\cite{wang2023mixed}& 0.951         & 0.968        & 0.966            & 0.949        & 0.971       & 0.976      & 0.964   \\
MixVoxels-X~\cite{wang2023mixed}& 0.954         & 0.968        & 0.971            & 0.953        & 0.973       & 0.976      & 0.966   \\
Im4D~\cite{lin2023im4d}& \NA           & \NA          & 0.970            & \NA          & \NA         & \NA        & 0.970   \\
4K4D~\cite{xu20234k4d}& \NA           & \NA          & 0.972            & \NA          & \NA         & \NA        & 0.972   \\ \midrule
\multicolumn{8}{c}{Dense COLMAP point cloud input}      \\
STG\textsuperscript{\textdaggerdbl}~\cite{li2023spacetime}& 0.949         & 0.974        & 0.976            & 0.950        & 0.980       & 0.981      & 0.968   \\
4DGS~\cite{yang2023real}& \NA           & \NA          & 0.980            & 0.960        & \NA         & \NA        & 0.972   \\ \midrule
\multicolumn{8}{c}{Sparse COLMAP point cloud input}      \\
STG\textsuperscript{\textdaggerdbl}~\cite{li2023spacetime}& \ocell 0.942 & \ycell 0.970 & \ocell 0.971 & \ocell 0.948 & \ocell 0.976 & \ocell 0.977 & \ocell 0.964   \\
4DGS~\cite{yang2023real}& \ycell 0.939 & \ocell 0.971 & \ycell 0.970 & \ycell 0.941 & \ycell 0.975 & \ycell 0.976 & \ycell 0.962   \\
4DGaussians~\cite{wu20234dgaussians}& 0.934        & 0.969        & 0.944        & 0.937        & 0.970        & 0.969        & 0.954   \\
\tbf{Ours}     & \rcell 0.951 & \rcell 0.976 & \rcell 0.977 & \rcell 0.956 & \rcell 0.980 & \rcell 0.979 & \rcell 0.970 \\
\bottomrule
\label{tab:N3V_SSIM2}
\end{tabular}%
}

\vspace{-1mm}

\resizebox{0.85\linewidth}{!}{%
\begin{tabular}{cccccccc}
\toprule
\multirow{2}{*}{Model\vspace{-15pt}} & \multicolumn{7}{c}{LPIPS} \\ \cmidrule(lr){2-8} 
 & \tworow{Coffee}{Martini} & \tworow{Cook}{Spinach} & \tworow{\!\!\!\!\!Cut\,Roasted\!\!\!\!\!}{Beef} & \tworow{Flame}{Salmon} & \tworow{Flame}{~Steak~} & \tworow{Sear}{~Steak~} & Average \\ \midrule
NeRFPlayer~\cite{song2023nerfplayer}& 0.085         & 0.113        & 0.144        & 0.098        & 0.088        & 0.138        & 0.111   \\
HyperReel~\cite{hyperreel}& 0.127         & 0.089        & 0.084        & 0.136        & 0.078        & 0.077        & 0.096   \\
Neural Volumes~\cite{lombardi2019neural}& \NA           & \NA          & \NA          & 0.295        & \NA          & \NA          & 0.295   \\
LLFF~\cite{mildenhall2019llff}& \NA           & \NA          & \NA          & 0.235        & \NA          & \NA          & 0.235   \\
DyNeRF~\cite{li2022neural}& \NA           & \NA          & \NA          & 0.083        & \NA          & \NA          & 0.083   \\
HexPlane~\cite{cao2023hexplane}& \NA           & 0.082        & 0.080        & 0.078        & 0.066        & 0.070        & 0.075   \\
MixVoxels-L~\cite{wang2023mixed}& 0.106         & 0.099        & 0.088        & 0.116        & 0.088        & 0.080        & 0.096   \\
MixVoxels-X~\cite{wang2023mixed}& 0.081         & 0.062        & 0.057        & 0.078        & 0.051        & 0.053        & 0.064   \\ \midrule
\multicolumn{8}{c}{Dense COLMAP point cloud input}      \\
STG\textsuperscript{\textdaggerdbl}~\cite{li2023spacetime}& 0.069         & 0.043        & 0.042        & 0.063        & 0.034        & 0.033        & 0.047   \\
4DGS~\cite{yang2023real}& \NA           & \NA          & 0.041        & \NA          & \NA          & \NA          & 0.055   \\ \midrule
\multicolumn{8}{c}{Sparse COLMAP point cloud input}      \\
STG\textsuperscript{\textdaggerdbl}~\cite{li2023spacetime}& \ycell 0.087  & \ycell 0.056 & \ycell 0.060 & \ocell 0.074 & \ycell 0.046 & \ycell 0.046 & \ycell 0.062   \\
4DGS~\cite{yang2023real}& \ocell 0.079  & \rcell 0.041 & \ocell 0.041 & \ycell 0.078 & \ocell 0.036 & \ocell 0.037 & \ocell 0.052 \\
4DGaussians~\cite{wu20234dgaussians}& 0.095         & \ycell 0.056 & 0.104        & 0.095        & 0.050        & \ycell 0.046 & 0.074   \\
\tbf{Ours}     & \rcell 0.070  & \ocell 0.042 & \rcell 0.040 & \rcell 0.066 & \rcell 0.034 & \rcell 0.035 & \rcell 0.048 \\
\bottomrule
\label{tab:N3V_LPIPS}
\end{tabular}%
}
\vspace{-4mm}
\end{table}

\clearpage 

{\small
\bibliographystyle{unsrt}
\bibliography{egbib}

\begin{thebibliography}{10}

\bibitem{mildenhall2020nerf}
B~Mildenhall, PP~Srinivasan, M~Tancik, JT~Barron, R~Ramamoorthi, and R~Ng.
\newblock Nerf: Representing scenes as neural radiance fields for view synthesis.
\newblock In {\em Proceedings of the European Conference on Computer Vision (ECCV)}, 2020.

\bibitem{pumarola2021d}
Albert Pumarola, Enric Corona, Gerard Pons-Moll, and Francesc Moreno-Noguer.
\newblock D-nerf: Neural radiance fields for dynamic scenes.
\newblock In {\em Proceedings of the IEEE/CVF Conference on Computer Vision and Pattern Recognition (CVPR)}, 2021.

\bibitem{park2021nerfies}
Keunhong Park, Utkarsh Sinha, Jonathan~T Barron, Sofien Bouaziz, Dan~B Goldman, Steven~M Seitz, and Ricardo Martin-Brualla.
\newblock Nerfies: Deformable neural radiance fields.
\newblock In {\em Proceedings of the IEEE/CVF Conference on Computer Vision and Pattern Recognition (CVPR)}, 2021.

\bibitem{li2021neural}
Zhengqi Li, Simon Niklaus, Noah Snavely, and Oliver Wang.
\newblock Neural scene flow fields for space-time view synthesis of dynamic scenes.
\newblock In {\em Proceedings of the IEEE/CVF Conference on Computer Vision and Pattern Recognition (CVPR)}, 2021.

\bibitem{gao2021dynamic}
Chen Gao, Ayush Saraf, Johannes Kopf, and Jia-Bin Huang.
\newblock Dynamic view synthesis from dynamic monocular video.
\newblock In {\em Proceedings of the IEEE/CVF International Conference on Computer Vision (ICCV)}, 2021.

\bibitem{park2021hypernerf}
Keunhong Park, Utkarsh Sinha, Peter Hedman, Jonathan~T Barron, Sofien Bouaziz, Dan~B Goldman, Ricardo Martin-Brualla, and Steven~M Seitz.
\newblock Hypernerf: a higher-dimensional representation for topologically varying neural radiance fields.
\newblock {\em ACM Transactions on Graphics (TOG)}, 2021.

\bibitem{li2022neural}
Tianye Li, Mira Slavcheva, Michael Zollhoefer, Simon Green, Christoph Lassner, Changil Kim, Tanner Schmidt, Steven Lovegrove, Michael Goesele, Richard Newcombe, et~al.
\newblock Neural 3d video synthesis from multi-view video.
\newblock In {\em Proceedings of the IEEE/CVF Conference on Computer Vision and Pattern Recognition (CVPR)}, 2022.

\bibitem{wu2022d}
Tianhao Wu, Fangcheng Zhong, Andrea Tagliasacchi, Forrester Cole, and Cengiz Oztireli.
\newblock D\^{} 2nerf: Self-supervised decoupling of dynamic and static objects from a monocular video.
\newblock In {\em Proceedings of the Neural Information Processing Systems (NeurIPS)}, 2022.

\bibitem{liu2022devrf}
Jia-Wei Liu, Yan-Pei Cao, Weijia Mao, Wenqiao Zhang, David~Junhao Zhang, Jussi Keppo, Ying Shan, Xiaohu Qie, and Mike~Zheng Shou.
\newblock Devrf: Fast deformable voxel radiance fields for dynamic scenes.
\newblock In {\em Proceedings of the Neural Information Processing Systems (NeurIPS)}, 2022.

\bibitem{li2022streaming}
Lingzhi Li, Zhen Shen, Zhongshu Wang, Li~Shen, and Ping Tan.
\newblock Streaming radiance fields for 3d video synthesis.
\newblock In {\em Proceedings of the Neural Information Processing Systems (NeurIPS)}, 2022.

\bibitem{cao2023hexplane}
Ang Cao and Justin Johnson.
\newblock Hexplane: A fast representation for dynamic scenes.
\newblock In {\em Proceedings of the IEEE/CVF Conference on Computer Vision and Pattern Recognition (CVPR)}, 2023.

\bibitem{fridovich2023k}
Sara Fridovich-Keil, Giacomo Meanti, Frederik~Rahb{\ae}k Warburg, Benjamin Recht, and Angjoo Kanazawa.
\newblock K-planes: Explicit radiance fields in space, time, and appearance.
\newblock In {\em Proceedings of the IEEE/CVF Conference on Computer Vision and Pattern Recognition (CVPR)}, 2023.

\bibitem{kerbl20233d}
Bernhard Kerbl, Georgios Kopanas, Thomas Leimk{\"u}hler, and George Drettakis.
\newblock 3d gaussian splatting for real-time radiance field rendering.
\newblock {\em ACM Transactions on Graphics (TOG)}, 2023.

\bibitem{wu20234dgaussians}
Guanjun Wu, Taoran Yi, Jiemin Fang, Lingxi Xie, Xiaopeng Zhang, Wei Wei, Wenyu Liu, Qi~Tian, and Wang Xinggang.
\newblock 4d gaussian splatting for real-time dynamic scene rendering.
\newblock In {\em Proceedings of the IEEE/CVF Conference on Computer Vision and Pattern Recognition (CVPR)}, 2024.

\bibitem{li2023spacetime}
Zhan Li, Zhang Chen, Zhong Li, and Yi~Xu.
\newblock Spacetime gaussian feature splatting for real-time dynamic view synthesis.
\newblock In {\em Proceedings of the IEEE/CVF Conference on Computer Vision and Pattern Recognition (CVPR)}, 2024.

\bibitem{zitnick2004high}
C~Lawrence Zitnick, Sing~Bing Kang, Matthew Uyttendaele, Simon Winder, and Richard Szeliski.
\newblock High-quality video view interpolation using a layered representation.
\newblock {\em ACM Transactions on Graphics (TOG)}, 2004.

\bibitem{bartels1995introduction}
Richard~H Bartels, John~C Beatty, and Brian~A Barsky.
\newblock {\em An introduction to splines for use in computer graphics and geometric modeling}.
\newblock Morgan Kaufmann, 1995.

\bibitem{Animating85ken}
Ken Shoemake.
\newblock Animating rotation with quaternion curves.
\newblock {\em Proceedings of the 12th annual conference on Computer graphics and interactive techniques}, 1985.

\bibitem{xu20234k4d}
Zhen Xu, Sida Peng, Haotong Lin, Guangzhao He, Jiaming Sun, Yujun Shen, Hujun Bao, and Xiaowei Zhou.
\newblock 4k4d: Real-time 4d view synthesis at 4k resolution.
\newblock In {\em Proceedings of the IEEE/CVF Conference on Computer Vision and Pattern Recognition (CVPR)}, 2024.

\bibitem{Sabater2017}
Neus Sabater, Guillaume Boisson, Benoit Vandame, Paul Kerbiriou, Frederic Babon, Matthieu Hog, Tristan Langlois, Remy Gendrot, Olivier Bureller, Arno Schubert, and Valerie Allie.
\newblock Dataset and pipeline for multi-view light-field video.
\newblock In {\em Proceedings of the IEEE/CVF Conference on Computer Vision and Pattern Recognition Workshop (CVPRW)}, 2017.

\bibitem{levoy1996light}
M~LEVOY.
\newblock Light field rendering.
\newblock In {\em Proceedings of ACM SIGGRAPH}, 1996.

\bibitem{buehler2001unstructured}
Chris Buehler, Michael Bosse, Leonard McMillan, Steven Gortler, and Michael Cohen.
\newblock Unstructured lumigraph rendering.
\newblock In {\em Proceedings of ACM SIGGRAPH}, 2001.

\bibitem{zhou2018stereo}
Tinghui Zhou, Richard Tucker, John Flynn, Graham Fyffe, and Noah Snavely.
\newblock Stereo magnification: Learning view synthesis using multiplane images.
\newblock In {\em Proceedings of ACM SIGGRAPH}, 2018.

\bibitem{flynn2019deepview}
John Flynn, Michael Broxton, Paul Debevec, Matthew DuVall, Graham Fyffe, Ryan Overbeck, Noah Snavely, and Richard Tucker.
\newblock Deepview: View synthesis with learned gradient descent.
\newblock In {\em Proceedings of the IEEE/CVF Conference on Computer Vision and Pattern Recognition (CVPR)}, 2019.

\bibitem{srinivasan2019pushing}
Pratul~P Srinivasan, Richard Tucker, Jonathan~T Barron, Ravi Ramamoorthi, Ren Ng, and Noah Snavely.
\newblock Pushing the boundaries of view extrapolation with multiplane images.
\newblock In {\em Proceedings of the IEEE/CVF Conference on Computer Vision and Pattern Recognition (CVPR)}, 2019.

\bibitem{kalantari2016learning}
Nima~Khademi Kalantari, Ting-Chun Wang, and Ravi Ramamoorthi.
\newblock Learning-based view synthesis for light field cameras.
\newblock In {\em Proceedings of ACM SIGGRAPH}, 2016.

\bibitem{meshry2019neural}
Moustafa Meshry, Dan~B Goldman, Sameh Khamis, Hugues Hoppe, Rohit Pandey, Noah Snavely, and Ricardo Martin-Brualla.
\newblock Neural rerendering in the wild.
\newblock In {\em Proceedings of the IEEE/CVF Conference on Computer Vision and Pattern Recognition (CVPR)}, 2019.

\bibitem{choi2019extreme}
Inchang Choi, Orazio Gallo, Alejandro Troccoli, Min~H Kim, and Jan Kautz.
\newblock Extreme view synthesis.
\newblock In {\em Proceedings of the IEEE/CVF International Conference on Computer Vision (ICCV)}, 2019.

\bibitem{xu2019deep}
Zexiang Xu, Sai Bi, Kalyan Sunkavalli, Sunil Hadap, Hao Su, and Ravi Ramamoorthi.
\newblock Deep view synthesis from sparse photometric images.
\newblock In {\em Proceedings of ACM SIGGRAPH}, 2019.

\bibitem{riegler2020free}
Gernot Riegler and Vladlen Koltun.
\newblock Free view synthesis.
\newblock In {\em Proceedings of the European Conference on Computer Vision (ECCV)}, 2020.

\bibitem{riegler2021stable}
Gernot Riegler and Vladlen Koltun.
\newblock Stable view synthesis.
\newblock In {\em Proceedings of the IEEE/CVF Conference on Computer Vision and Pattern Recognition (CVPR)}, 2021.

\bibitem{hu2021worldsheet}
Ronghang Hu, Nikhila Ravi, Alexander~C Berg, and Deepak Pathak.
\newblock Worldsheet: Wrapping the world in a 3d sheet for view synthesis from a single image.
\newblock In {\em Proceedings of the IEEE/CVF International Conference on Computer Vision (ICCV)}, 2021.

\bibitem{chen2019learning}
Wenzheng Chen, Huan Ling, Jun Gao, Edward Smith, Jaakko Lehtinen, Alec Jacobson, and Sanja Fidler.
\newblock Learning to predict 3d objects with an interpolation-based differentiable renderer.
\newblock In {\em Proceedings of the Neural Information Processing Systems (NeurIPS)}, 2019.

\bibitem{genova2018unsupervised}
Kyle Genova, Forrester Cole, Aaron Maschinot, Aaron Sarna, Daniel Vlasic, and William~T Freeman.
\newblock Unsupervised training for 3d morphable model regression.
\newblock In {\em Proceedings of the IEEE/CVF Conference on Computer Vision and Pattern Recognition (CVPR)}, 2018.

\bibitem{liu2019soft}
Shichen Liu, Tianye Li, Weikai Chen, and Hao Li.
\newblock Soft rasterizer: A differentiable renderer for image-based 3d reasoning.
\newblock In {\em Proceedings of the IEEE/CVF International Conference on Computer Vision (ICCV)}, 2019.

\bibitem{yu2021pixelnerf}
Alex Yu, Vickie Ye, Matthew Tancik, and Angjoo Kanazawa.
\newblock pixelnerf: Neural radiance fields from one or few images.
\newblock In {\em Proceedings of the IEEE/CVF Conference on Computer Vision and Pattern Recognition (CVPR)}, 2021.

\bibitem{kim2022infonerf}
Mijeong Kim, Seonguk Seo, and Bohyung Han.
\newblock Infonerf: Ray entropy minimization for few-shot neural volume rendering.
\newblock In {\em Proceedings of the IEEE/CVF Conference on Computer Vision and Pattern Recognition (CVPR)}, 2022.

\bibitem{yang2023freenerf}
Jiawei Yang, Marco Pavone, and Yue Wang.
\newblock Freenerf: Improving few-shot neural rendering with free frequency regularization.
\newblock In {\em Proceedings of the IEEE/CVF Conference on Computer Vision and Pattern Recognition (CVPR)}, 2023.

\bibitem{jain2021putting}
Ajay Jain, Matthew Tancik, and Pieter Abbeel.
\newblock Putting nerf on a diet: Semantically consistent few-shot view synthesis.
\newblock In {\em Proceedings of the IEEE/CVF International Conference on Computer Vision (ICCV)}, 2021.

\bibitem{tancik2022block}
Matthew Tancik, Vincent Casser, Xinchen Yan, Sabeek Pradhan, Ben Mildenhall, Pratul~P Srinivasan, Jonathan~T Barron, and Henrik Kretzschmar.
\newblock Block-nerf: Scalable large scene neural view synthesis.
\newblock In {\em Proceedings of the IEEE/CVF Conference on Computer Vision and Pattern Recognition (CVPR)}, 2022.

\bibitem{turki2022mega}
Haithem Turki, Deva Ramanan, and Mahadev Satyanarayanan.
\newblock Mega-nerf: Scalable construction of large-scale nerfs for virtual fly-throughs.
\newblock In {\em Proceedings of the IEEE/CVF Conference on Computer Vision and Pattern Recognition (CVPR)}, 2022.

\bibitem{barron2022mipnerf360}
Jonathan~T. Barron, Ben Mildenhall, Dor Verbin, Pratul~P. Srinivasan, and Peter Hedman.
\newblock Mip-nerf 360: Unbounded anti-aliased neural radiance fields.
\newblock {\em CVPR}, 2022.

\bibitem{kaizhang2020}
Kai Zhang, Gernot Riegler, Noah Snavely, and Vladlen Koltun.
\newblock Nerf++: Analyzing and improving neural radiance fields.
\newblock {\em arXiv:2010.07492}, 2020.

\bibitem{leegeometry}
Junoh Lee, Hyunjun Jung, Jin-Hwi Park, Inhwan Bae, and Hae-Gon Jeon.
\newblock Geometry-aware projective mapping for unbounded neural radiance fields.
\newblock In {\em Proceedings of the International Conference on Learning Representations (ICLR)}, 2024.

\bibitem{wang2023f2}
Peng Wang, Yuan Liu, Zhaoxi Chen, Lingjie Liu, Ziwei Liu, Taku Komura, Christian Theobalt, and Wenping Wang.
\newblock F2-nerf: Fast neural radiance field training with free camera trajectories.
\newblock In {\em Proceedings of the IEEE/CVF Conference on Computer Vision and Pattern Recognition (CVPR)}, 2023.

\bibitem{yang2022neumesh}
Bangbang Yang, Chong Bao, Junyi Zeng, Hujun Bao, Yinda Zhang, Zhaopeng Cui, and Guofeng Zhang.
\newblock Neumesh: Learning disentangled neural mesh-based implicit field for geometry and texture editing.
\newblock In {\em Proceedings of the European Conference on Computer Vision (ECCV)}, 2022.

\bibitem{kellnhofer2021neural}
Petr Kellnhofer, Lars~C Jebe, Andrew Jones, Ryan Spicer, Kari Pulli, and Gordon Wetzstein.
\newblock Neural lumigraph rendering.
\newblock In {\em Proceedings of the IEEE/CVF Conference on Computer Vision and Pattern Recognition (CVPR)}, 2021.

\bibitem{yariv2020multiview}
Lior Yariv, Yoni Kasten, Dror Moran, Meirav Galun, Matan Atzmon, Basri Ronen, and Yaron Lipman.
\newblock Multiview neural surface reconstruction by disentangling geometry and appearance.
\newblock In {\em Proceedings of the Neural Information Processing Systems (NeurIPS)}, 2020.

\bibitem{zhang2021nerfactor}
Xiuming Zhang, Pratul~P Srinivasan, Boyang Deng, Paul Debevec, William~T Freeman, and Jonathan~T Barron.
\newblock Nerfactor: Neural factorization of shape and reflectance under an unknown illumination.
\newblock {\em ACM Transactions on Graphics (TOG)}, 2021.

\bibitem{zhang2021physg}
Kai Zhang, Fujun Luan, Qianqian Wang, Kavita Bala, and Noah Snavely.
\newblock Physg: Inverse rendering with spherical gaussians for physics-based material editing and relighting.
\newblock In {\em Proceedings of the IEEE/CVF Conference on Computer Vision and Pattern Recognition (CVPR)}, 2021.

\bibitem{yang2022ps}
Wenqi Yang, Guanying Chen, Chaofeng Chen, Zhenfang Chen, and Kwan-Yee~K Wong.
\newblock Ps-nerf: Neural inverse rendering for multi-view photometric stereo.
\newblock In {\em Proceedings of the European Conference on Computer Vision (ECCV)}, 2022.

\bibitem{boss2022samurai}
Mark Boss, Andreas Engelhardt, Abhishek Kar, Yuanzhen Li, Deqing Sun, Jonathan Barron, Hendrik Lensch, and Varun Jampani.
\newblock Samurai: Shape and material from unconstrained real-world arbitrary image collections.
\newblock In {\em Proceedings of the Neural Information Processing Systems (NeurIPS)}, 2022.

\bibitem{ye2023intrinsicnerf}
Weicai Ye, Shuo Chen, Chong Bao, Hujun Bao, Marc Pollefeys, Zhaopeng Cui, and Guofeng Zhang.
\newblock Intrinsicnerf: Learning intrinsic neural radiance fields for editable novel view synthesis.
\newblock In {\em Proceedings of the IEEE/CVF International Conference on Computer Vision (ICCV)}, 2023.

\bibitem{sitzmann2021light}
Vincent Sitzmann, Semon Rezchikov, Bill Freeman, Josh Tenenbaum, and Fredo Durand.
\newblock Light field networks: Neural scene representations with single-evaluation rendering.
\newblock In {\em Proceedings of the Neural Information Processing Systems (NeurIPS)}, 2021.

\bibitem{wang2022r2l}
Huan Wang, Jian Ren, Zeng Huang, Kyle Olszewski, Menglei Chai, Yun Fu, and Sergey Tulyakov.
\newblock R2l: Distilling neural radiance field to neural light field for efficient novel view synthesis.
\newblock In {\em Proceedings of the European Conference on Computer Vision (ECCV)}, 2022.

\bibitem{yu2021plenoctrees}
Alex Yu, Ruilong Li, Matthew Tancik, Hao Li, Ren Ng, and Angjoo Kanazawa.
\newblock Plenoctrees for real-time rendering of neural radiance fields.
\newblock In {\em Proceedings of the IEEE/CVF International Conference on Computer Vision (ICCV)}, 2021.

\bibitem{chen2022tensorf}
Anpei Chen, Zexiang Xu, Andreas Geiger, Jingyi Yu, and Hao Su.
\newblock Tensorf: Tensorial radiance fields.
\newblock In {\em Proceedings of the European Conference on Computer Vision (ECCV)}, 2022.

\bibitem{adelson1991plenoptic}
Edward~H Adelson, James~R Bergen, et~al.
\newblock {\em The plenoptic function and the elements of early vision}, volume~2.
\newblock Vision and Modeling Group, Media Laboratory, Massachusetts Institute of Technology, 1991.

\bibitem{weng2022humannerf}
Chung-Yi Weng, Brian Curless, Pratul~P Srinivasan, Jonathan~T Barron, and Ira Kemelmacher-Shlizerman.
\newblock Humannerf: Free-viewpoint rendering of moving people from monocular video.
\newblock In {\em Proceedings of the IEEE/CVF Conference on Computer Vision and Pattern Recognition (CVPR)}, 2022.

\bibitem{xu2021h}
Hongyi Xu, Thiemo Alldieck, and Cristian Sminchisescu.
\newblock H-nerf: Neural radiance fields for rendering and temporal reconstruction of humans in motion.
\newblock In {\em Proceedings of the Neural Information Processing Systems (NeurIPS)}, 2021.

\bibitem{alldieck2021imghum}
Thiemo Alldieck, Hongyi Xu, and Cristian Sminchisescu.
\newblock imghum: Implicit generative models of 3d human shape and articulated pose.
\newblock In {\em Proceedings of the IEEE/CVF International Conference on Computer Vision (ICCV)}, 2021.

\bibitem{loper2015smpl}
Matthew Loper, Naureen Mahmood, Javier Romero, Gerard Pons-Moll, and Michael~J Black.
\newblock Smpl: A skinned multi-person linear model.
\newblock {\em ACM Transactions on Graphics (TOG)}, 2015.

\bibitem{jiang2023instantavatar}
Tianjian Jiang, Xu~Chen, Jie Song, and Otmar Hilliges.
\newblock Instantavatar: Learning avatars from monocular video in 60 seconds.
\newblock In {\em Proceedings of the IEEE/CVF Conference on Computer Vision and Pattern Recognition (CVPR)}, 2023.

\bibitem{athar2022rignerf}
ShahRukh Athar, Zexiang Xu, Kalyan Sunkavalli, Eli Shechtman, and Zhixin Shu.
\newblock Rignerf: Fully controllable neural 3d portraits.
\newblock In {\em Proceedings of the IEEE/CVF Conference on Computer Vision and Pattern Recognition (CVPR)}, 2022.

\bibitem{bai2023high}
Yunpeng Bai, Yanbo Fan, Xuan Wang, Yong Zhang, Jingxiang Sun, Chun Yuan, and Ying Shan.
\newblock High-fidelity facial avatar reconstruction from monocular video with generative priors.
\newblock In {\em Proceedings of the IEEE/CVF Conference on Computer Vision and Pattern Recognition (CVPR)}, 2023.

\bibitem{peng2021animatable}
Sida Peng, Junting Dong, Qianqian Wang, Shangzhan Zhang, Qing Shuai, Xiaowei Zhou, and Hujun Bao.
\newblock Animatable neural radiance fields for modeling dynamic human bodies.
\newblock In {\em Proceedings of the IEEE/CVF International Conference on Computer Vision (ICCV)}, 2021.

\bibitem{xian2021space}
Wenqi Xian, Jia-Bin Huang, Johannes Kopf, and Changil Kim.
\newblock Space-time neural irradiance fields for free-viewpoint video.
\newblock In {\em Proceedings of the IEEE/CVF Conference on Computer Vision and Pattern Recognition (CVPR)}, 2021.

\bibitem{zhang2021editable}
Jiakai Zhang, Xinhang Liu, Xinyi Ye, Fuqiang Zhao, Yanshun Zhang, Minye Wu, Yingliang Zhang, Lan Xu, and Jingyi Yu.
\newblock Editable free-viewpoint video using a layered neural representation.
\newblock {\em ACM Transactions on Graphics (TOG)}, 2021.

\bibitem{tschernezki2021neuraldiff}
Vadim Tschernezki, Diane Larlus, and Andrea Vedaldi.
\newblock Neuraldiff: Segmenting 3d objects that move in egocentric videos.
\newblock In {\em 2021 International Conference on 3D Vision (3DV)}, 2021.

\bibitem{li2023dynibar}
Zhengqi Li, Qianqian Wang, Forrester Cole, Richard Tucker, and Noah Snavely.
\newblock Dynibar: Neural dynamic image-based rendering.
\newblock In {\em Proceedings of the IEEE/CVF Conference on Computer Vision and Pattern Recognition (CVPR)}, 2023.

\bibitem{jiang2022neuman}
Wei Jiang, Kwang~Moo Yi, Golnoosh Samei, Oncel Tuzel, and Anurag Ranjan.
\newblock Neuman: Neural human radiance field from a single video.
\newblock In {\em Proceedings of the European Conference on Computer Vision (ECCV)}, 2022.

\bibitem{song2023nerfplayer}
Liangchen Song, Anpei Chen, Zhong Li, Zhang Chen, Lele Chen, Junsong Yuan, Yi~Xu, and Andreas Geiger.
\newblock Nerfplayer: A streamable dynamic scene representation with decomposed neural radiance fields.
\newblock {\em IEEE Transactions on Visualization and Computer Graphics}, 2023.

\bibitem{luiten2023dynamic}
Jonathon Luiten, Georgios Kopanas, Bastian Leibe, and Deva Ramanan.
\newblock Dynamic 3d gaussians: Tracking by persistent dynamic view synthesis.
\newblock In {\em Proceedings of the International Conference on 3D Vision}, 2024.

\bibitem{yang2023real}
Zeyu Yang, Hongye Yang, Zijie Pan, and Li~Zhang.
\newblock Real-time photorealistic dynamic scene representation and rendering with 4d gaussian splatting.
\newblock In {\em Proceedings of the International Conference on Learning Representations (ICLR)}, 2023.

\bibitem{yang2023deformable}
Ziyi Yang, Xinyu Gao, Wen Zhou, Shaohui Jiao, Yuqing Zhang, and Xiaogang Jin.
\newblock Deformable 3d gaussians for high-fidelity monocular dynamic scene reconstruction.
\newblock In {\em Proceedings of the IEEE/CVF Conference on Computer Vision and Pattern Recognition (CVPR)}, 2024.

\bibitem{huang2023sc}
Yi-Hua Huang, Yang-Tian Sun, Ziyi Yang, Xiaoyang Lyu, Yan-Pei Cao, and Xiaojuan Qi.
\newblock Sc-gs: Sparse-controlled gaussian splatting for editable dynamic scenes.
\newblock In {\em Proceedings of the IEEE/CVF Conference on Computer Vision and Pattern Recognition (CVPR)}, 2024.

\bibitem{qian20233dgsavatar}
Zhiyin Qian, Shaofei Wang, Marko Mihajlovic, Andreas Geiger, and Siyu Tang.
\newblock 3dgs-avatar: Animatable avatars via deformable 3d gaussian splatting.
\newblock In {\em Proceedings of the IEEE/CVF Conference on Computer Vision and Pattern Recognition (CVPR)}, 2024.

\bibitem{hu2023gauhuman}
Shoukang Hu and Ziwei Liu.
\newblock Gauhuman: Articulated gaussian splatting from monocular human videos.
\newblock In {\em Proceedings of the IEEE/CVF Conference on Computer Vision and Pattern Recognition (CVPR)}, 2024.

\bibitem{lu20243d}
Zhicheng Lu, Xiang Guo, Le~Hui, Tianrui Chen, Min Yang, Xiao Tang, Feng Zhu, and Yuchao Dai.
\newblock 3d geometry-aware deformable gaussian splatting for dynamic view synthesis.
\newblock In {\em Proceedings of the IEEE/CVF Conference on Computer Vision and Pattern Recognition (CVPR)}, 2024.

\bibitem{lin2023gaussian}
Youtian Lin, Zuozhuo Dai, Siyu Zhu, and Yao Yao.
\newblock Gaussian-flow: 4d reconstruction with dynamic 3d gaussian particle.
\newblock In {\em Proceedings of the IEEE/CVF Conference on Computer Vision and Pattern Recognition (CVPR)}, 2024.

\bibitem{fridovich2022plenoxels}
Sara Fridovich-Keil, Alex Yu, Matthew Tancik, Qinhong Chen, Benjamin Recht, and Angjoo Kanazawa.
\newblock Plenoxels: Radiance fields without neural networks.
\newblock In {\em Proceedings of the IEEE/CVF Conference on Computer Vision and Pattern Recognition (CVPR)}, 2022.

\bibitem{chen2023gaussianeditor}
Yiwen Chen, Zilong Chen, Chi Zhang, Feng Wang, Xiaofeng Yang, Yikai Wang, Zhongang Cai, Lei Yang, Huaping Liu, and Guosheng Lin.
\newblock Gaussianeditor: Swift and controllable 3d editing with gaussian splatting.
\newblock In {\em Proceedings of the IEEE/CVF Conference on Computer Vision and Pattern Recognition (CVPR)}, 2024.

\bibitem{Yu2024MipSplatting}
Zehao Yu, Anpei Chen, Binbin Huang, Torsten Sattler, and Andreas Geiger.
\newblock Mip-splatting: Alias-free 3d gaussian splatting.
\newblock In {\em Proceedings of the IEEE/CVF Conference on Computer Vision and Pattern Recognition (CVPR)}, 2024.

\bibitem{schonberger2016structure}
Johannes~L Schonberger and Jan-Michael Frahm.
\newblock Structure-from-motion revisited.
\newblock In {\em Proceedings of the IEEE/CVF Conference on Computer Vision and Pattern Recognition (CVPR)}, 2016.

\bibitem{liu2019radam}
Liyuan Liu, Haoming Jiang, Pengcheng He, Weizhu Chen, Xiaodong Liu, Jianfeng Gao, and Jiawei Han.
\newblock On the variance of the adaptive learning rate and beyond.
\newblock In {\em Proceedings of the International Conference on Learning Representations (ICLR)}, 2020.

\bibitem{hyperreel}
Benjamin Attal, Jia-Bin Huang, Christian Richardt, Michael Zollhoefer, Johannes Kopf, Matthew O’Toole, and Changil Kim.
\newblock Hyperreel: High-fidelity 6-dof video with ray-conditioned sampling.
\newblock In {\em Proceedings of the IEEE/CVF Conference on Computer Vision and Pattern Recognition (CVPR)}, 2023.

\bibitem{lombardi2019neural}
Stephen Lombardi, Tomas Simon, Jason Saragih, Gabriel Schwartz, Andreas Lehrmann, and Yaser Sheikh.
\newblock Neural volumes: learning dynamic renderable volumes from images.
\newblock {\em ACM Transactions on Graphics (TOG)}, 2019.

\bibitem{mildenhall2019llff}
Ben Mildenhall, Pratul~P. Srinivasan, Rodrigo Ortiz-Cayon, Nima~Khademi Kalantari, Ravi Ramamoorthi, Ren Ng, and Abhishek Kar.
\newblock Local light field fusion: Practical view synthesis with prescriptive sampling guidelines.
\newblock {\em ACM Transactions on Graphics (TOG)}, 2019.

\bibitem{wang2023mixed}
Feng Wang, Sinan Tan, Xinghang Li, Zeyue Tian, Yafei Song, and Huaping Liu.
\newblock Mixed neural voxels for fast multi-view video synthesis.
\newblock In {\em Proceedings of the IEEE/CVF International Conference on Computer Vision (ICCV)}, 2023.

\bibitem{lin2023im4d}
Haotong Lin, Sida Peng, Zhen Xu, Tao Xie, Xingyi He, Hujun Bao, and Xiaowei Zhou.
\newblock High-fidelity and real-time novel view synthesis for dynamic scenes.
\newblock In {\em Proceedings of SIGGRAPH Asia}, 2023.

\bibitem{sun20243dgstream}
Jiakai Sun, Han Jiao, Guangyuan Li, Zhanjie Zhang, Lei Zhao, and Wei Xing.
\newblock 3dgstream: On-the-fly training of 3d gaussians for efficient streaming of photo-realistic free-viewpoint videos.
\newblock In {\em Proceedings of the IEEE/CVF Conference on Computer Vision and Pattern Recognition (CVPR)}, 2024.

\end{thebibliography}
}

\end{document}